\documentclass[10pt]{article} %
\usepackage[preprint]{tmlr}

\usepackage{amsmath,amsfonts,bm}

\def\eqref#1{equation~\ref{#1}}

\def\1{\bm{1}}

\def\eps{{\epsilon}}

\DeclareMathAlphabet{\mathsfit}{\encodingdefault}{\sfdefault}{m}{sl}
\SetMathAlphabet{\mathsfit}{bold}{\encodingdefault}{\sfdefault}{bx}{n}

\DeclareMathOperator*{\argmax}{arg\,max}

\usepackage{microtype}
\usepackage{diagbox,booktabs} %
\usepackage{xcolor}
\usepackage{url}
\usepackage{amsmath}
\usepackage{amssymb}
\usepackage{mathtools}
\usepackage{amsthm}
\definecolor{citecolor}{HTML}{0071bc}
\usepackage[breaklinks=true,bookmarks=false,colorlinks,bookmarks=false, citecolor=citecolor, pagebackref]{hyperref}
\usepackage[capitalize,noabbrev]{cleveref}
\usepackage{url}
\usepackage[margin=3cm]{geometry}
\usepackage{algorithm}
\let\classAND\AND
\let\AND\relax
\usepackage{algorithmic}

\let\AND\classAND
\AtBeginEnvironment{algorithmic}{\let\AND\algoAND}
\usepackage{xspace}
\usepackage{graphicx}
\usepackage{bm}
\usepackage[inline]{enumitem}

\usepackage{tabularx}
\usepackage{array}
\usepackage{makecell}
\usepackage{multirow}
\usepackage{multicol}
\usepackage[font=small,labelfont=bf]{caption}
\usepackage{subcaption}
\usepackage{colortbl}
\usepackage{color}
\let\oldAND\AND %
\renewcommand{\AND}{\oldAND} %
\makeatletter
\DeclareRobustCommand\onedot{\futurelet\@let@token\@onedot}
\def\@onedot{\ifx\@let@token.\else.\null\fi\xspace}
\makeatother
\definecolor{Blue}{rgb}{0.88,1,1}
\definecolor{Gray}{gray}{0.85}
\definecolor{customgreen}{rgb}{0.196,0.871,0.518}
\definecolor{customred}{rgb}{0.992,0.361,0.388}

\newcommand*{\belowrulesepcolor}[1]{%
  \noalign{%
    \kern-\belowrulesep
    \begingroup
      \color{#1}%
      \hrule height\belowrulesep
    \endgroup
  }%
}
\newcommand*{\aboverulesepcolor}[1]{%
  \noalign{%
    \begingroup
      \color{#1}%
      \hrule height\aboverulesep
    \endgroup
    \kern-\aboverulesep
  }%
}
\title{SelfEval: Leveraging discriminative nature of generative models for evaluation}

\author{\name Sai Saketh Rambhatla \email rssaketh@meta.com \\
      \addr GenAI, Meta
      \AND
      \name Ishan Misra \email imisra@meta.com \\
      \addr GenAI, Meta
}

\newcommand{\supplementarysection}{%
  \setcounter{figure}{0}%
  \let\oldthefigure\thefigure%
  \setcounter{section}{0}
  \let\oldthesection\thesection%
  \setcounter{table}{0}
  \let\oldthetable\thetable%

}
\newcommand{\supplementarytitle}{
  \title{
    Appendix
}

\maketitle
\supplementarysection
}

\newcommand{\OURS}{\textsc{SelfEval}\xspace}
\newcommand{\Ours}{\OURS}
\newcommand{\DATASET}{Internal-Dataset\xspace}
\newcommand{\taskStyle}[1]{\texttt{#1}}

\def\eg{\emph{e.g}\onedot} 
\def\ie{\emph{i.e}\onedot} 
 
\def\etc{\emph{etc}\onedot}

\newcommand{\bx}{\mathbf{x}}
\newcommand{\bc}{\mathbf{c}}
\newcommand{\bmu}{\bm{\mu}}
\newcommand{\bSigma}{\mathbf{\Sigma}}
\newcommand{\bigO}{\mathcal{O}}
\begin{document}

\maketitle

\begin{abstract}
    We present an automated way to evaluate the text alignment of text-to-image generative diffusion models using standard image-text recognition datasets.
    Our method, called \OURS, uses the generative model to compute the likelihood of real images given text prompts, and the likelihood can be used to perform recognition tasks with the generative model.
    We evaluate generative models on standard datasets created for multimodal text-image discriminative learning and assess fine-grained aspects of their performance: attribute binding, color recognition, counting, shape recognition, spatial understanding.
    Existing automated metrics rely on an external pretrained model like CLIP (VLMs) or LLMs, and are sensitive to the exact pretrained model and its limitations.
    \OURS sidesteps these issues, and to the best of our knowledge, is the first automated metric to show a high degree of agreement for measuring text-faithfulness with the gold-standard human evaluations across multiple generative models, benchmarks and evaluation metrics.
    \OURS also reveals that generative models showcase competitive recognition performance on challenging tasks such as Winoground image-score compared to discriminative models.
    We hope \OURS enables easy and reliable automated evaluation for diffusion models.
\end{abstract}

\section{Introduction}
\label{sec:intro}

In the past few years, generative image models have rapidly advanced and state-of-the-art text-to-image models now generate high quality realistic images.
While a lot of research effort is focussed on improving these models, their evaluation has received considerably less attention.
Evaluations for text-to-image models typically focus on two aspects: (1) quality of the generated image; and (2) the alignment between the generated image and the input text, \ie, the `faithfulness' of the generation.
The gold standard for evaluating text-to-image models is to compare generations from pairs of models using human judgement.
However, using pairwise human evaluations does not scale to lots of models or generations, making it difficult to convert them to ordinal metrics to rank models.
Thus, automatic evaluations are commonly used as a proxy for comparing models.

\begin{figure}
  \centering
  \includegraphics[width=\textwidth]{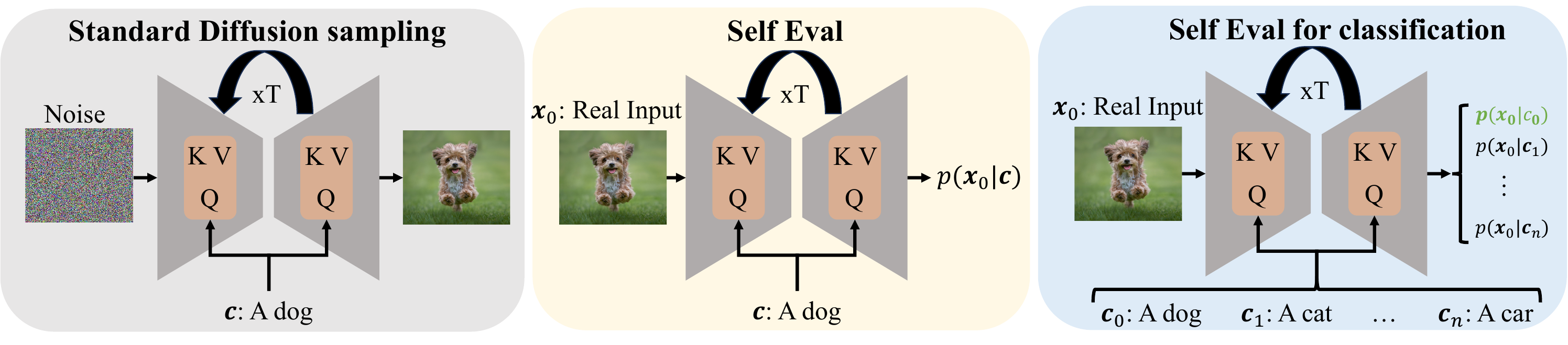}
\caption{\textbf{Illustration of proposed method:} (Left) Starting from a noised input, the standard diffusion sampling method denoises the input iteratively to generate images from the input distribution.
(Middle): \OURS takes a pair (image $\bx_0$ and conditioning $\bc$) to estimate the likelihood $p(\bx_0 | \bc)$ of the pair in an iterative fashion. (Right): Given an image, $\bx_0$ and $n$ captions, $\{\bc_0, \bc_1, \dotsc, \bc_n\}$, \OURS is a principled way to convert generative models into discriminative models.
We show that the classification performance of these classifiers can be used to evaluate the generative capabilities.}
\label{fig:main_fig}
\end{figure}

In this work, we focus on automatic evaluations that measure the text adhering capabilities of a generative diffusion model and ask the question: can the diffusion model itself be used to measure the relatedness of an image-text pair and thus evaluate its own generations?
Most works on text-to-image diffusion models focus on sampling good images given a text prompt.
However, as shown in Figure~\ref{fig:main_fig}, diffusion models can be used to estimate the conditional likelihood of an image $\bx$ given a text prompt $\bc$, \ie, $p(\bx | \bc)$.
We propose \OURS which is a practical way to estimate such likelihoods accounting for numerical issues arising in standard diffusion models.
We show that these likelihoods can be used directly to solve recognition tasks and evaluate the model's text-faithfulness ability.

\OURS repurposes standard multimodal image-text datasets such as Visual Genome, COCO and CLEVR to measure the model's text understanding capabilities.
\OURS uses \textbf{ground truth} (real) image-text pairs for evaluation and computes classification accuracy making it more robust and interpretable.
Our evaluation allows us to assess fine-grained aspects such as the model's ability to recognize colors, count objects \etc.
We apply our method to a wide variety of diffusion models: different types of image representations (pixel based, latent space based), different text encoders and model sizes.
\OURS's automatic evaluation results are in agreement with the `gold-standard' human judgements making \OURS suitable for evaluation.

\begin{figure*}
    \centering
    \includegraphics[width=\textwidth]{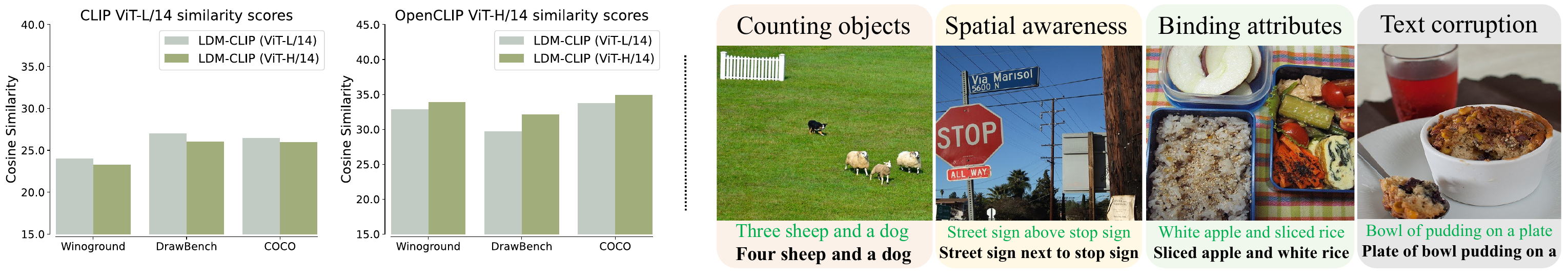}
  \caption{\textbf{Drawbacks of CLIP for generative model evaluation}. (Left) We compare the CLIP similarity scores of two Latent diffusion models~\cite{rombach2022high} trained with CLIP ViT-L/14 (LDM-CLIP (ViT-L/14)) and OpenCLIP ViT-H/14 (LDM-CLIP (ViT-H/14)) text encoders.
  On the left, we compare the CLIP similarity scores, computed using CLIP ViT-L/14, on prompts generated from DrawBench, Winoground and, COCO datasets.
  The plot on the right compares the CLIP similarity scores computed using OpenCLIP ViT-H/14 model. The ranking changes depending on the model used. (Right) CLIP has poor performance in tasks involving counting instances, spatial relationships, matching attributes to objects and understanding corruption of text
  which constitute about 50 (25\%) prompts in DrawBench. In each example, the correct caption is shown in green and CLIP picked the caption in bold. Using CLIP to evaluate text to image models on such prompts is not optimal.
   }\label{fig:clip-drawbacks}
  \end{figure*}

Existing automated evaluations for text faithfulness of generative models rely on an additional external discriminative model, \eg, CLIP or LLMs, to measure the `relatedness' of the generated image to the input text.
As we show in~\cref{fig:clip-drawbacks}, relying on an external model leads to three major issues.
First, the automated metrics vary greatly depending on the type of the external model used for evaluation and they often have an arbitrary range, e.g., See Table~\ref{exp:tab:comp_metrics_pdm},\ref{exp:tab:comp_metrics_ldm} for the range of CLIP~\cite{Radford2021LearningTV} and MID~\cite{kim2022mid} scores.
Second, many generative models rely on an external model such as CLIP's text encoding during training, and thus using the same CLIP model for automated evaluation biases the results.
Finally, the external model itself can have several limitations (like poor performance on few image-text tasks in the case of CLIP and hallucination in the case of LLMs~\cite{Xu2024HallucinationII}) making its scores unreliable.

\OURS only uses the generative model and thus its scores directly reflect the strengths and weaknesses of the generative model.
Note that most automated metrics operate on generated images whereas \OURS uses real images from image-text recognition datasets.

\section{Related Works}
\label{sec:related}
\noindent\underline{\textbf{Generative models}}: Generative models learn to model the joint distribution, $p(X,Y)$ of data consisting of an observed variable $X$ and the target $Y$.
The model can then be used to sample novel data.
In this work, we are interested in image generation models \ie, models that learn the distribution of natural images.

Generative Adverserial Networks (GAN)~\cite{Goodfellow2014GenerativeAN,Radford2015UnsupervisedRL}, Variational AutoEncoders (VAE)~\cite{KingmaW13} and Denoising Diffusion Probabilistic models (DDPM)~\cite{HoJA20} are some of the most popular image generation models in the literature. GANs belong to the category of generative models, where two distinct components, a generator and a discriminator,
 are pitted against each other within a zero-sum game framework.
 VAEs are a category of autoencoders that ensure ``regularity'' within the latent space by constraining their distribution to closely align with a well-behaved and typically standard normal distribution.
 In more recent times, DDPMs have exceeded the capabilities of all preceding state-of-the-art image generative models in terms of their generative prowess.
Drawing inspiration from non-equilibrium statistical physics, Diffusion probabilistic models~\cite{pmlr-v37-sohl-dickstein15} employ a forward diffusion process to gradually destroy the structure in the unknown input distribution and transforming it into a well-behaved and tractable distribution.
A reverse diffusion process is trained to learn to restore the structure, thereby learning the input distribution.
An explicit connection between diffusion models and denoising score matching is established in~\cite{HoJA20}, leading to a simplified objective for training diffusion models.
We utilize diffusion models in this study due to their outstanding image generation performance, as demonstrated in~\cite{Dhariwal2021DiffusionMB}.

\noindent\underline{\textbf{Diffusion models}}:
In a relatively short time, diffusion models have surpassed GANs and VAEs as the defacto models for image generation due to their superior quality~\cite{Dhariwal2021DiffusionMB} and flexibility.
Numerous studies have shown that diffusion models can be conditioned on a variety of modalities, including object classes~\cite{Peebles2022DiT,HoJA20}, natural language captions~\cite{Saharia2022PhotorealisticTD,rombach2022high,NicholDRSMMSC22,ramesh2022hierarchical}, camera pose~\cite{liu2023zero1to3}, images~\cite{brooks2022instructpix2pix}, bounding boxes~\cite{Li2023GLIGENOG}, segmentation, edge and depth maps~\cite{zhang2023adding}.
Among these, text-conditioned diffusion models have attracted significant interest and popularity.
Given paired image, caption data, text conditioned diffusion models are trained to fuse the caption features, extracted using a pre-trained text encoder, with the image features using cross attention.
Text-to-Image models demonstrate a remarkable comprehension of compositionality within text, often highlighted by their capacity to generate images based on counterfactual textual descriptions (like avacado shaped chair \etc).
The most popular text encoders in use today for text-conditioned image synthesis are text encoders from the CLIP~\cite{Radford2021LearningTV} and the text-to-text transformer T5~\cite{2020t5}.
In this work, we analyze the text understanding capabilities of the diffusion models trained with different text encoders.

There exist two families of diffusion models in the literature, namely, pixel~\cite{Saharia2022PhotorealisticTD,ramesh2022hierarchical} and latent diffusion~\cite{rombach2022high}, differing primarily in the nature of input.
The diffusion process in pixel diffusion is performed on pixels making these models computationally expensive.
Latent diffusion models~\cite{rombach2022high} operate on the autoencoder's latent space, balancing the computational constraints with the quality and flexibility of pixel diffusion models.
In this work, we analyze the text understanding capabilities of two state-of-the-art models with different text encoders each from pixel and latent diffusion models.

\noindent\underline{\textbf{Classifiers with diffusion models}}:
Lately, there has been a increase in the usage of conditional diffusion models as classifiers, driven by their superior understanding of the conditioned modality.
These models are surprisingly good at capturing intricate patterns and dependencies within the conditioning input, making them strong discriminative models across a range of downstream tasks.
Notable works include~\cite{he2023discriminative}, \cite{mukhopadhyay2023diffusion} that either finetune a diffusion model, or use linear probing, for several classification and reasoning tasks.
Unlike these methods, we do not train any models but instead convert the generative model into a discriminative one to understand its text understanding capabilities.
Along similar lines to \Ours, \cite{clark2023texttoimage, li2023diffusion} employ the ELBO loss as a proxy to estimate the likelihood scores (and subsequently the posterior using Bayes' rule) from diffusion models for several image understanding tasks.
\cite{li2023diffusion}also report promising results on ITM tasks on the Winoground~\cite{Thrush2022WinogroundPV} dataset.
Instead, we propose a systematic way, accounting for numerical stability, to estimate the likelihood of an image given the text from a conditioned diffusion model.
To the best of our knowledge, \Ours is the first method to show that the discriminative performance of generative models aligns closely with 'gold-standard' human evaluations in the assessment of generative models.

\noindent\underline{\textbf{Image-Text Matching for evaluating generative models}}:
CLIP R-precision~\cite{park2021benchmark} is a metric to evaluate the text understanding capabilities of generative models similar to \Ours.
While CLIP R-precision measures the text retrieval performance given a \textbf{generated} image, \OURS uses the \textbf{ground truth} image-text pairs for evaluation.
Using generated images poses a problem since the generated images are often out-of-distribution to the external model, \ie CLIP, making the score unreliable.
\Ours avoids this issue by using the generative model, instead of an external model, for evaluation.

\section{Method: Converting generative models to discriminative models}\label{sec:method}

Our method converts generative (diffusion) models into discriminative models by simply changing the inference, and does not require any retraining.
This allows us to use the diffusion model itself on a variety of different image-text benchmarks and assess the diffusion model's image-text understanding capabilities.
We briefly discuss an overview of diffusion models in Sec.~\ref{method_subsec:prelim} followed by our proposed method in Sec.~\ref{method_subsec:method}

\subsection{Preliminaries}\label{method_subsec:prelim}
Diffusion Probabilistic Models (DPM)~\cite{pmlr-v37-sohl-dickstein15} belong to a class of generative models trained to `denoise' inputs created by a Markovian \textit{forward} process.
The forward process starts with a sample $\bx_0$ and repeatedly adds Gaussian noise over $t$ timesteps to generate $\bx_t$:
\begin{align}
    q(\mathbf{x}_t | \mathbf{x}_{t-1}) \sim \mathcal{N}(\mathbf{x}_{t}; \sqrt{1-\beta_t}\mathbf{x}_{t-1}, \beta_t\mathbf{I}).
\end{align}

Here $q(\bx_0)$ is the data distribution. $\beta_t$ is the strength of the noise at timestep $t$ with $\beta_0 = 0, \beta_T=1$.
Note that $\bx_t$ are the same size as the input.
The joint distribution of the input along with the latents $q(\bx_{0:T})$ is
\begin{align}q(\bx_{0:T}) = q(\bx_0) \prod_{t=1}^T q(\bx_t|\bx_{t-1})\end{align}

To sample images, one applies the \emph{reverse} process, $p(\bx_{t-1} | \bx_t)\allowbreak$, starting with $\bx_T$ sampled from the unit normal distribution, $\mathcal{N}(\mathbf{0}, \mathbb{I})\allowbreak$.
So the joint distribution of the reverse process can be described as
\begin{align}p(\bx_{0:T}) = p(\bx_T) \prod_{t=1}^{T}p(\bx_{t-1} | \bx_t)\end{align}

The reverse process $p(\bx_{t-1} | \bx_t)$ is not tractable and is often modeled using a neural network whose parameters are characterized by $\theta$, \ie $p_{\theta}(\bx_{t-1} | \bx_t) \sim \mathcal{N}(\bx_{t-1}; \bmu_\theta(\bx_t, t), \bSigma_\theta(\bx_t,t))$.

\subsection{Likelihood estimates from diffusion models}\label{method_subsec:method}
We specifically focus on text-to-image diffusion models, although our formulation extends to any conditional diffusion model.
Text-to-image diffusion models are trained on a large datasets of image-text $(\bx, \bc)$ pairs and model the reverse diffusion process $p(\bx_{t-1} | \bx_t, \bc)$.
We `invert' such a generative model and use it to estimate the likelihood of a real image $\bx$ given a text caption $\bc$, \ie, $p(\bx | \bc)$.
We note that our method only changes the inference of a diffusion model and does not require any training.
Assuming uniform prior on the classes, the likelihood $p(\bx|\bc)$ can be converted into the posterior, $p(\bc|\bx)$ using Bayes' rule, \ie $p(\bc | \bx) = \frac{p(\bx | \bc)}{|\mathcal{C}|}$, where $\mathcal{C}$ is the set of all classes.

Given the reverse process of a diffusion model parameterized by $\theta$, the likelihood for a datapoint $\bx_0$ is
\begin{align}
    p_{\theta}(\bx_0 | \bc) &= \int{p_{\theta}(\bx_{0:T} | \bc)d\bx_{1:T}}  \\
 &= \int p(\bx_T)\prod_{t=1}^T p_{\theta}(\bx_{t-1}| \bx_t, \bc) d\bx_{1:T} \label{eq:rep}
\end{align}
Since the diffusion models reverse process $p_{\theta}(\cdot)$ is also a gaussian, we can further write this as

\begin{align}
    p(\bx_0|\bc)
    &= \int p(\bx_T)\prod_{t=1}^T \frac{1}{\sqrt{(2\pi)^D} |\Sigma_\theta|} \nonumber \\
    &\quad \times \text{exp}\Bigg(-\frac{1}{2}(\bx_{t-1} - \bmu_\theta(\bx_t,\bc))^T \times \bSigma_\theta^{-1}(\bx_{t-1} - \bmu_\theta(\bx_t,\bc))\Bigg)d\bx_{1:T}\label{eq:est}
    \end{align}

Here, $p(\bx_T) \sim \mathcal{N}(0,\mathbb{I})$.
For the sake of simplicity, we denote any realization of the random variable $\bx_0$ as $\bx_0$. Given a natural language caption $\bc$, an image $\bx_0$ and the noised latents $x_{1:T}$, the quantity inside the integral in Eq.~\ref*{eq:est} can be estimated numerically. We compute a Monte Carlo estimate of the integral by sampling
$N$ noise terms ($\eps$) and computing $p(\bx_0|\bc)$ as

\begin{align}
    p(\bx_0|\bc) &= \sum_{n=1}^{N} p(\bx^n_T)\prod_{t=1}^T p(\bx^n_{t-1} | \bx^n_t, \bc) \nonumber \\
    &\quad \text{where } \bx_t^n = \sqrt{1-\beta_t}\bx_{t-1}^n + \sqrt{\beta_{t}}\eps^n   \label{eq:final_og}
\end{align}
   \begin{figure*}[!t]
    \includegraphics[width=\linewidth]{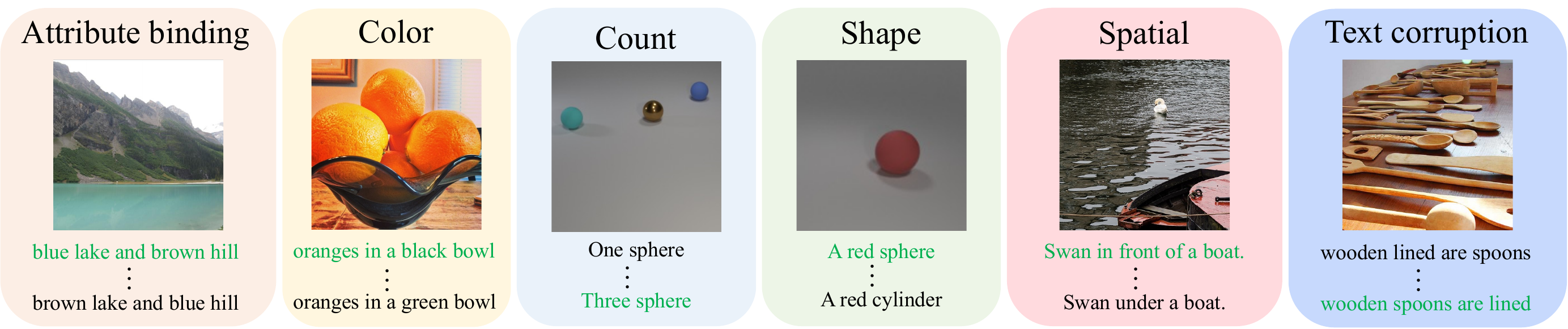}
    \caption{\textbf{Representative samples from the benchmark.} We divide the evaluation into six broad tasks, namely \taskStyle{Attribute binding}, \taskStyle{Color}, \taskStyle{Count}, \taskStyle{Shape}, \taskStyle{Spatial}, and \taskStyle{Text Corruption}.
    Each task is designed to evaluate a specific aspect of text faithfulness mimicing the categories in DrawBench. Each task is posed as an image-text matching problem, where given an image, the goal is to pick the right caption among distractors.
    The figure above shows examples from each task with the right caption highlighted in green.}\label{exp:fig:data_fig}
\end{figure*}

\noindent{\textbf{Practical considerations.}}
The terms on the RHS of Eq.~\ref{eq:final_og} are multivariate gaussians and analytically computing them involves exponentials which can be numerically unstable.
This can be prevented by computing log probabilities instead. Taking log both sides of Eq.~\ref{eq:final_og}, we get

\begin{align}
 \log{p(\bx_0 | \bc)} &= \log{\sum_{n=1}^{N} p(\bx^n_T)\prod_{t=1}^T p(\bx^n_{t-1} | \bx^n_t, \bc)} \\
 &\geq \sum_{n=1}^{N} (\log{p(\bx^n_T)} +  \sum_{t=1}^T \log{p(\bx^n_{t-1} | \bx^n_t, \bc)}) \label{eq:final_sim}
\end{align}
Where Eq.~\ref*{eq:final_sim} is from Jensen's inequality for concave functions, \ie $\mathbb{E}(f(x)) \leq f(\mathbb{E}(x))$.
All the terms in Eq.~\ref*{eq:final_sim} are log probabilities of multivariate gaussians, which can be computed analytically and are numerically more stable.

We now show how estimating $p(\bx_0 | \bc)$ allows us to use a diffusion model for discriminative tasks and thus to evaluate their image-text understanding capabilities.

\subsection{\OURS to evaluate diffusion model's text faithfulness}

The text faithfulness of a diffusion model measures its ability to understand the text prompt and ground it in the generated image output.
The `standard' way of evaluating text faithfulness uses a manually curated list of text prompts to generate images.
The `alignment' between the generated images and the text prompts can be measured using an external model or a human evaluator.
The text faithfulness of a generative model inherently also measures its vision-language reasoning abilities.
Thus, in \OURS, we propose to directly measure the generative model's vision-language discriminative performance as a way to evaluate its text faithfulness.

We pose the \OURS evaluation as an image-text matching problem and measure the generative model's recognition performance on standard discriminative image-text datasets.
Thus, \OURS does not rely on external models such as CLIP, does not need human evaluators, and does not need manual text prompt-set curation.

Image-text matching problems such as image-classification or retrieval can be reformulated as picking the correct caption for a single image $\bx$ from a set of captions $\{\bc_i\}$.
We can use a diffusion model to estimate $p(\bx | \bc_i)$ for each of the captions and pick the caption that gives the highest likelihood.
As shown in Fig.~\ref{fig:main_fig}, the noised latents $\bx_{1:T}$ are computed using the forward process. The final latent $\bx_T$ is
denoised for $T$ steps using the reverse process to obtain the denoised latents $\bar{\bx}_{0:T-1}$. This process is repeated for $N$ independent noise vectors resulting in $\{\bx_{1:T}^n\}_{n=1}^N$, $\{\bar{\bx}_{0:T-1}^n\}_{n=1}^N$. Next,
the likelihood can be computed as $p(\bx_0|c_k) = \sum_{n=1}^N p(\bx_T^n)\prod_{t=1}^T p(\bx_{t-1}^n | \bar{\bx}_t^n, \bc_k)$, which is then converted to the posterior, $p(\bc_k | \bx_0)$ using Bayes' rule. Finally, the caption with the highest likelihood, \ie. $\argmax_{\bc_i} p(\bc_i|\bx_0)$ is chosen as the right one.

\section{Experiments}
\label{sec:experiments}
\begin{figure*}[!t]
    \centering

    \begin{subfigure}[b]{\textwidth}

        \centering
        \includegraphics[width=0.24\linewidth]{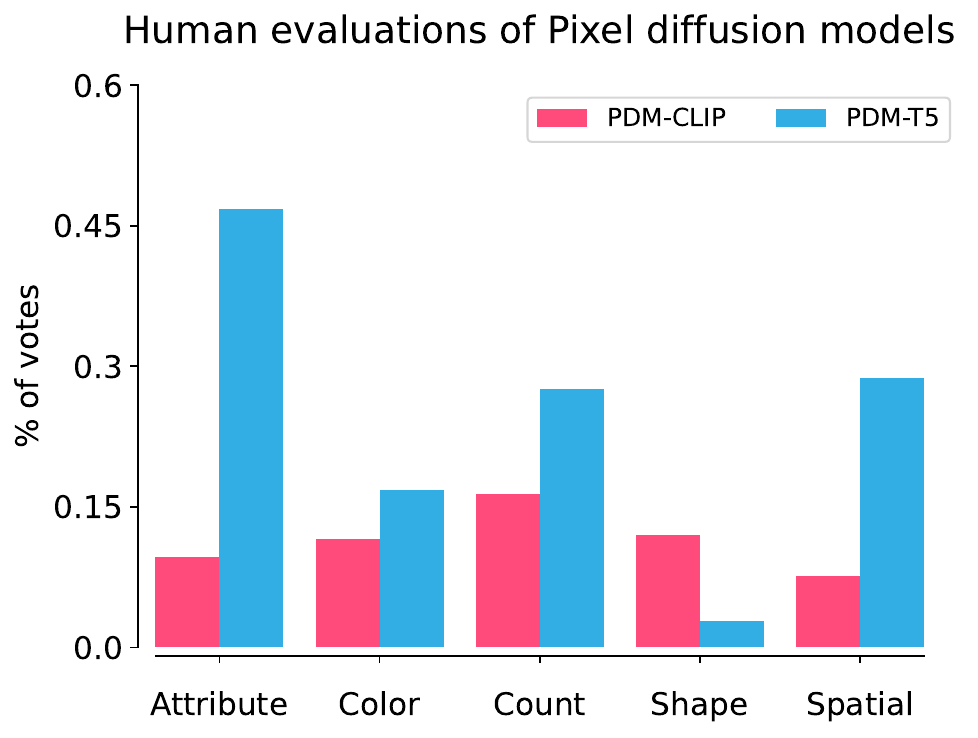}%
        \hfill
        \includegraphics[width=0.24\linewidth]{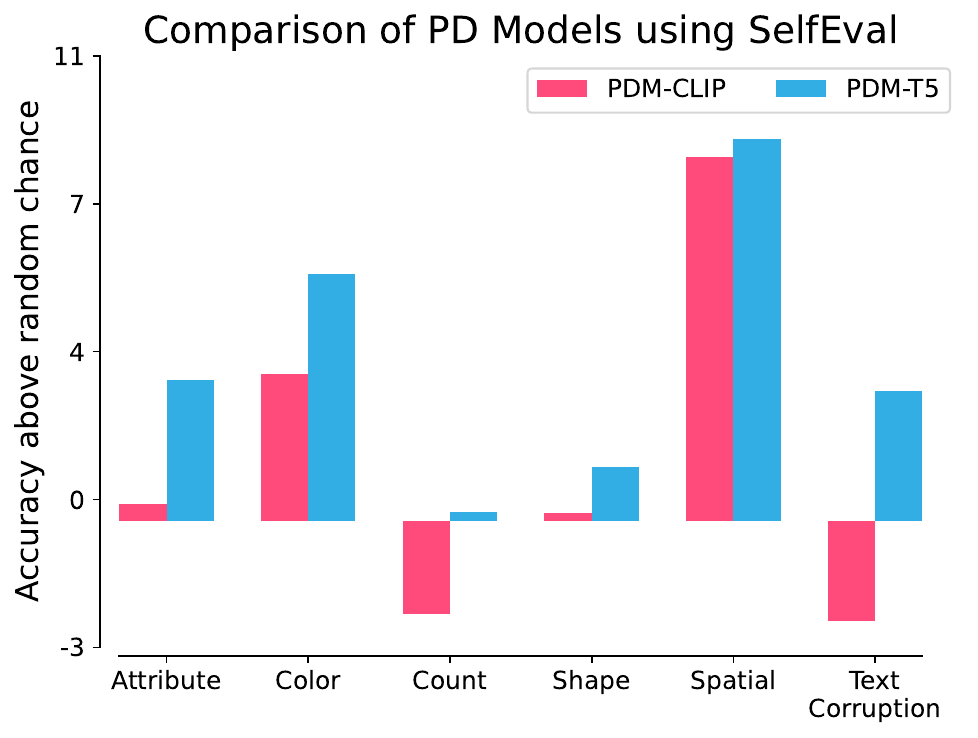}%
        \label{exp:fig:main_res_pdm}
        \centering
        \includegraphics[width=0.24\linewidth]{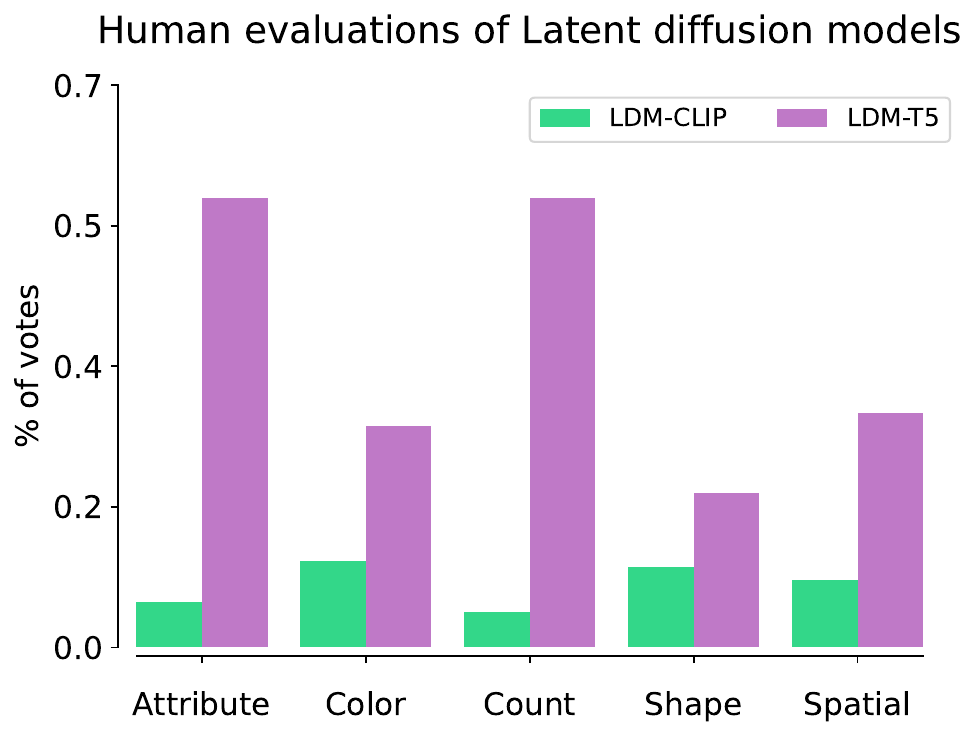}%
        \hfill
        \includegraphics[width=0.24\linewidth]{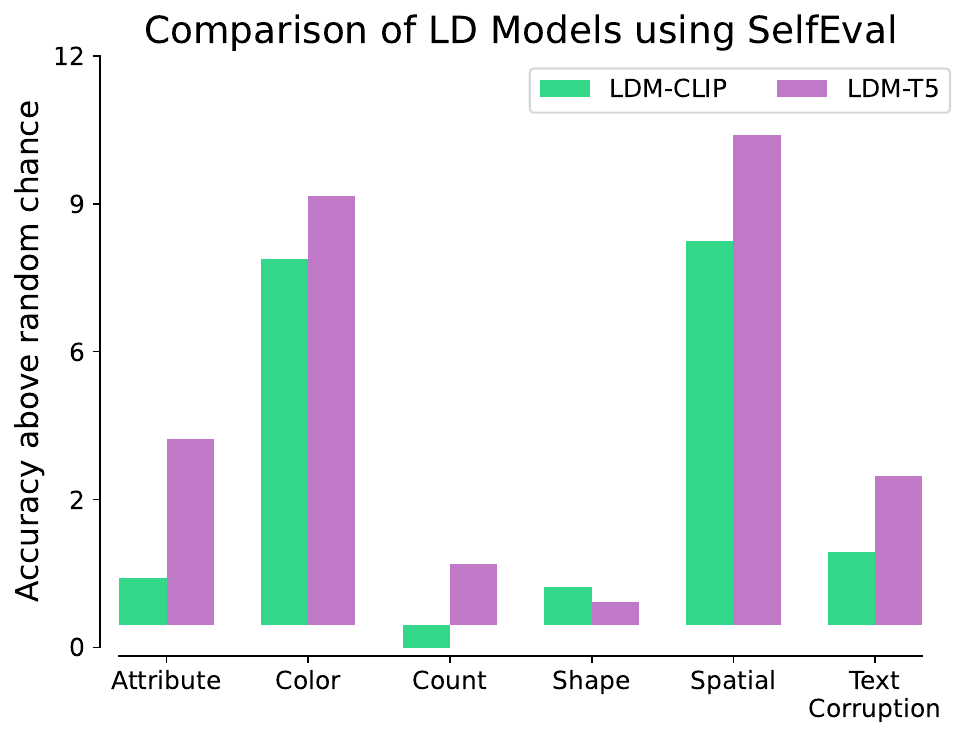}%

        \label{exp:fig:main_res_ldm}
    \end{subfigure}
    \caption{\textbf{Evaluating text-to-image models} using human evaluations and \OURS. We evaluate different types of text-to-image models such as pixel diffusion (first two columns) and latent diffusion model (last two columns), and models that use different text encoders such as T5 XXL and CLIP. We observe that across all 4 diffusion models the relative ordering given by \OURS's accuracy correlates with the pairwise human evaluation results. We also observe that latent diffusion models have a higher \OURS accuracy than pixel diffusion models suggesting better text-faithfulness.
    Using the stronger T5 text encoder leads to better performance across human evaluations and \OURS.
    }\label{exp:fig:main_res_all}
\end{figure*}
We now use \OURS to evaluate text-to-image diffusion models.
In~\ref{subsec:exp:benchmark}, we introduce our benchmark datasets and models, and present the \OURS results in~\cref{subsec:exp:main_exp}.

\subsection{Benchmark and Evaluation}\label{subsec:exp:benchmark}
In \OURS, we pose the text faithfulness evaluation as an image-text matching task, where the goal is to pick the right image caption pair among distractors.
\par \noindent \textbf{Tasks.}
We identify six broad reasoning tasks for evaluation (\cref{exp:fig:data_fig}): 1) \taskStyle{Attribute binding},
 2) \taskStyle{Color}, 3) \taskStyle{Count}, 4) \taskStyle{Shape}, 5) \taskStyle{Spatial relationships}, and 6) \taskStyle{Text corruption}.
Each of these tasks evaluate the model's understanding of a specific aspect of text faithfulness and is similar to the categories of prompts from DrawBench~\cite{Saharia2022PhotorealisticTD}.
The six tasks are constructed using data from TIFA~\cite{hu2023tifa}, CLEVR~\cite{Johnson2016CLEVRAD} and ARO~\cite{yuksekgonul2023when}.
\par \noindent \textbf{Datasets.}
\noindent \textbf{TIFA}~\cite{hu2023tifa} consists of 4000 text prompts, collected manually and from image captioning datasets, to evaluate the text faithfulness of generative models.
In our evaluation, we use $\sim$2000 of these text-prompts that are constructed from the COCO~\cite{2014-lin} dataset and convert the dataset from question-answering to an image-text matching format as detailed in the supplement.
\noindent \textbf{Attribution, Relation and Order (ARO)}~\cite{yuksekgonul2023when} is a benchmark that uses data from Visual Genome~\cite{vg} for attribute and spatial relations, and COCO for ordering tasks.
\noindent \textbf{CLEVR}~\cite{Johnson2016CLEVRAD} is a benchmark for compositional understanding and visual reasoning using synthetic images.
We adopt the splits proposed by~\cite{Lewis2022DoesCB} for our case.

\noindent
We divide the datasets among all the reasoning task as follows.
For attribute binding, we combine samples from ARO (attribution) and CLEVR.
For colors and counts, we use corresponding samples from TIFA and CLEVR.
For shapes, we use samples from CLEVR.
Data for spatial relationships is from TIFA, CLEVR and ARO (relations).
The data for the text corruption task is from the ARO (order sensitivity) dataset.
A sample of each task consists of an image and multiple text prompts and the performance on the task is the accuracy of pairing the image with the right caption.

We measure the performance of text-to-image generative models on the benchmark using the following evaluation methods.

\par \noindent \textbf{\OURS (Ours)} is an automatic evaluation method and uses both the images and text from our benchmark introduced in~\cref{subsec:exp:benchmark}.
For each benchmark task, we randomly sample $1000$ examples and evaluate the classification performance on them. We repeat this three times and the report the mean accuracy.
We use 10 trials (\ie$N=10$) and perform diffusion for $100$ steps (\ie$T=100$) for all the models. Refer to the supplement for ablation experiments on $N$, $T$.

\par \noindent \textbf{Human evaluations} are the gold standard for judging the performance of text-to-image models using pairwise comparsions.
We present humans with generations from two models and ask them to vote for one of four choices: ``both'' the generations are faithful, ``none'' of them are faithful, or if only one of the two images (``Image 1'' or ``Image 2'') demonstrates fidelity to the given prompt.
For simplicity, we only report votes where there is a clear preference for a model.
We randomly pick $250$ text prompts from each benchmark task as conditioning for human evaluation and the images are generated using DDIM~\cite{song2021denoising} sampling, with $100$ denoising steps.
Note that unlike \OURS, human evaluations do \emph{not} use the real images from the benchmark tasks and the human evaluators only look at the generated images.

\subsubsection{Models}\label{subsubsec:exp:benchmark:models}
We use models with different image representations: pixel diffusion models which directly use the pixel RGB values, and latent diffusion models where the image is projected into a latent space using an auto-encoder.
We pick models trained with different text encoders within each class.
This enables us to analyze the effect of text encoder on the final performance within each class.

\par \noindent{\textbf{Diffusion models with CLIP text encoder.}}

We employ a model trained with the OpenCLIP~\cite{ilharco_gabriel_2021_5143773} text encoder with a ViT-H/14 backbone for latent diffusion, accessed via an API containing open-sourced model weights.
This model, trained on a public dataset of 5 billion images (excluding explicit material), outputs images of $512\times512$ resolution.
For pixel diffusion, we use the architecture of DALL-E-2~\cite{ramesh2022hierarchical} in our experiments and train a model.
This model uses a CLIP (ViT-L/14) text encoder, produces images of $64\times 64$ resolution, and has a total of 4.2B parameters.
It is trained for 2M steps on an internal image-text dataset (\DATASET).

\noindent{\textbf{Diffusion models with T5 text encoder.}}
We train a UNet model for latent diffusion, similar to~\cite{rombach2022high}, but with the CLIP text encoder replaced by a T5 XXL~\cite{2020t5} text encoder.
This model outputs images of $256\times256$ resolution.
Trained on \DATASET for 2M steps using a latent space with a $4\times$ downsampling factor, the model has a total of 5.8B parameters.
We train a pixel diffusion model with 7.5B parameters, similar to Imagen~\cite{Saharia2022PhotorealisticTD}, on $64\times64$ resolution inputs for 2M steps using the same data.
Following this, we use a super-resolution model to upsample the output to $512\times 512$.
With the exception of the CLIP-based latent diffusion model~\cite{rombach2022high}, all the other models are trained for the same number of steps on the exact same data to ensure fair comparison.
\subsection{Main results}\label{subsec:exp:main_exp}

We evaluate the four text-to-image models and report results in~\cref{exp:fig:main_res_all}.
For \OURS, we report the accuracy difference with the random chance accuracy, since each of the tasks has a different degree of difficulty.

\par \noindent \textbf{Agreement between \OURS and human evaluation.}
In~\cref{exp:fig:main_res_all}, we evaluate four different diffusion models using both human evaluation and \OURS.
The human evaluation performance, measured via pairwise comparison, aligns with the ranking given by \OURS for both pixel and latent diffusion models.
To our knowledge, this is the first work to correlate the discriminative performance of generative models with human evaluation for text-to-image diffusion models across various models and tasks.
The strong alignment between \OURS and human raters suggests that \OURS is a reliable and interpretable way for evaluating and comparing the text faithfulness of different diffusion models.

Next, we use \OURS to further analyze the performance of diffusion models.

\par \noindent \textbf{Effect of the text encoder.}
Comparing the different text-encoders used in~\cref{exp:fig:main_res_all}, we observe that diffusion models using the stronger T5 text encoder perform better on most tasks than the ones using the CLIP text encoder.
The stronger performance of T5-based models holds for both human evaluations and \OURS.
The \OURS results indicate that diffusion models using the CLIP-based encoders perform poorly on the \taskStyle{Count} task, even worse than random chance.
For the \taskStyle{Text Corruption} task, which involves identifying a linguistically correct sentence among distractors with shuffled word order, the performance of CLIP-based models is lower.
Thus, as suggested by prior work~\cite{yuksekgonul2023when}, CLIP models exhibit a bag-of-words understanding of the text.

\par \noindent \textbf{Pixel vs latent diffusion.}
We compare the \OURS performance of the pixel diffusion models to that of the latent diffusion models in~\cref{exp:fig:main_res_ldmvpdm}.
Among models that use the same text encoder, \ie PDM-T5 and LDM-T5, we observe that the latent diffusion models outperform the pixel diffusion ones in most cases, especially on the harder tasks of \taskStyle{Attribute Binding}, \taskStyle{Count}, \taskStyle{Spatial Relations} and \taskStyle{Text Corruption}.
We hypothesize that this difference can be explained by the fact that the latent diffusion models operate on the compressed latent space and prioritize the text conditioning while `offloading' the high-frequency image details to the autoencoder.
We further investigate the performance of pixel and latent diffusion models by employing human raters to evaluate their text faithfulness in~\cref{exp:fig:main_res_ldmvpdm}.
The data, for human evaluation, is constructed by randomly picking $500$ examples from the all the tasks ($100$ examples from each task except text corruption), and choosing the right caption as the text prompt.
We convert the accuracy of \OURS, to votes, by counting the number of samples where only one model is right.
From~\cref{exp:fig:main_res_ldmvpdm}, we observe that human raters prefer the generations of latent diffusion models over pixel diffusion models for text faithfulness.
\OURS also shows that latent diffusion models have a better text faithfulness and shows an alignment with human evaluations.

\begin{table}[!t]
    \begin{minipage}[t]{0.41\linewidth}
        \centering
        \vspace{0pt}
        \setlength\abovecaptionskip{0pt}
        \includegraphics[width=\textwidth]{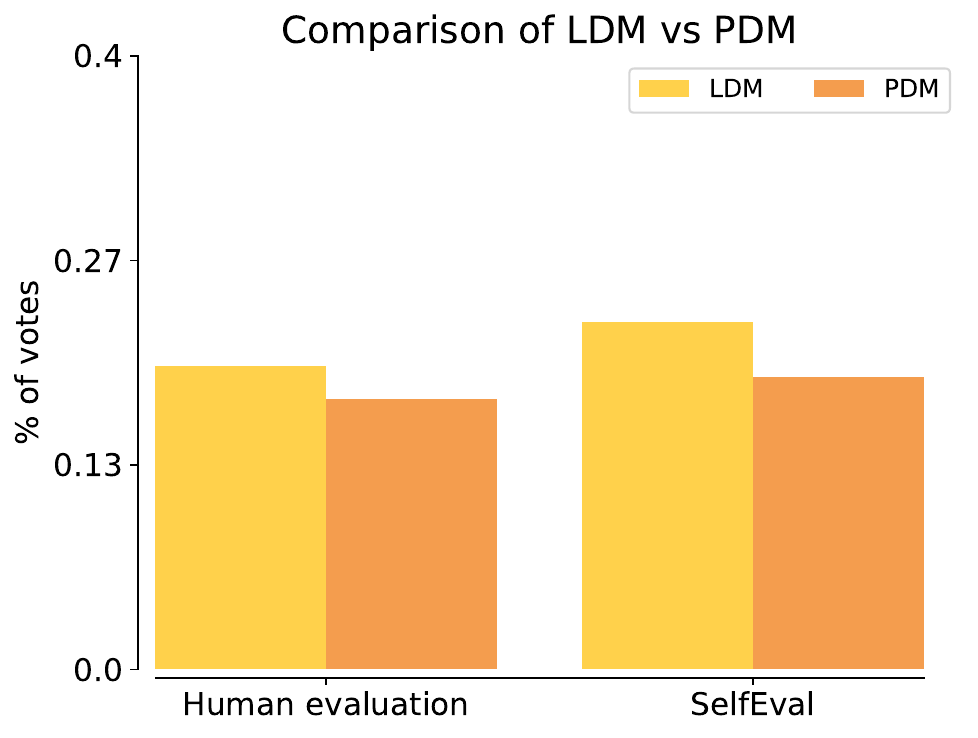}
        \captionof{figure}{\textbf{Pixel vs Latent diffusion.} Human raters rank the generation of latent models higher than pixel models in text faithfulness. \OURS exhibits a similar trend. }\label{exp:fig:main_res_ldmvpdm}
    \end{minipage}\hfill
    \begin{minipage}[t]{0.54\linewidth}
        \renewcommand{\arraystretch}{1.05}
        \setlength{\tabcolsep}{8pt}
        \centering
        \footnotesize
        \caption{\textbf{Diffusion models evaluated on the Winoground dataset}. We measure the image score (accuracy of picking the correct image given a text prompt) and text score (accuracy of picking the correct text given an image). Using \OURS allows us to use diffusion models for both tasks unlike~\cite{li2023diffusion} which leads to zero image score.}
        \label{exp:tab:add:retr}
        \resizebox{\linewidth}{!}{
        \begin{tabular}{@{}lccc@{}}
        \toprule
        Method & Model & Image Score & Text Score \\
         \midrule
        CLIP (ViT-L/14) & $-$ & 8.00 & 30.25\\
        OCLIP (ViT-H/14) & $-$& 12.75 & 30.75\\
        \midrule
        ~\cite{li2023diffusion} & LDM-CLIP & 0 & 34.00\\
        \OURS & LDM-CLIP & 7.25 & 22.75  \\
        \midrule
        \OURS & PDM-CLIP & 14.00 &  17.00 \\
         \OURS& PDM-T5 & 12.00 & 28.25 \\
         \OURS& LDM-T5 & 13.50 & 29.00 \\
        \bottomrule
        \end{tabular}
        }
    \end{minipage}\hfill

\end{table}

\begin{table}[!t]
    \begin{minipage}[t]{\linewidth}
        \renewcommand{\arraystretch}{1.05}
        \setlength{\tabcolsep}{8pt}
        \centering
        \footnotesize
        \caption{\textbf{Comparison of evaluation metrics on pixel diffusion models.} We compare \Ours with existing metrics on pixel diffusion models. "Humans" represents the winning votes obtained by the models in the human evaluation. Each row uses a green or red cell to denote whether the ranking agrees or disagrees with human judgment respectively. Note that the value ranges for CLIPScore and MID differ and are not directly comparable to those of the other metrics..}
        \label{exp:tab:comp_metrics_pdm}
        \newcolumntype{R}{c<{\hspace*{-\tabcolsep}}}
        \resizebox{\linewidth}{!}{
        \begin{tabular}{@{}lcccccccccR@{}}
        \toprule
        Method & \multicolumn{2}{c}{Attribute} & \multicolumn{2}{c}{Color} & \multicolumn{2}{c}{Count} &  \multicolumn{2}{c}{Shape} & \multicolumn{2}{c}{Spatial} \\
         & \multicolumn{2}{c}{binding} &  &  &   &  \\
        \midrule
         & CLIP & T5  & CLIP  & T5 & CLIP & T5  & CLIP  & T5 & CLIP & T5  \\

         \midrule
        Humans & 24 & \cellcolor{customgreen}117 & 29 & \cellcolor{customgreen}42 & 41 & \cellcolor{customgreen}69 & \cellcolor{customgreen}30 & 7 & 19 & \multicolumn{1}{r}{\cellcolor{customgreen}72} \\
        CLIPScore ($\uparrow$) & 0.90 & \cellcolor{customgreen}0.98 & 0.89 & \cellcolor{customgreen}0.91 & \cellcolor{customred}0.81 & 0.76 & \cellcolor{customgreen}0.86 & 0.82& 0.76 & \multicolumn{1}{r}{\cellcolor{customgreen}0.78}\\
        MID ($\downarrow$) & -8.6$\mathrm{E}$14 & \cellcolor{customgreen}-2.1$\mathrm{E}$14& -2.7$\mathrm{E}$5 & \cellcolor{customgreen}-1.8$\mathrm{E}$5 & -8.6$\mathrm{E}$3 & \cellcolor{customgreen}-1.1$\mathrm{E}$4 & -4.6$\mathrm{E}$3 & \cellcolor{customred}-2.6$\mathrm{E}$3 & \cellcolor{customred}-1.0$\mathrm{E}$15 & \multicolumn{1}{r}{-5.6$\mathrm{E}$15}\\
        VPEval ($\uparrow$)&  61.6 & \cellcolor{customgreen}63.1 & \cellcolor{customred}86.9 & 86.0 & 16.7 & \cellcolor{customgreen}31.9 & \cellcolor{customgreen}94.6 & 91.9 & 18.1 & \multicolumn{1}{r}{\cellcolor{customgreen}25.3}\\
        LLMScore ($\uparrow$)&  6.6 & \cellcolor{customgreen}8.2 & 15.5 & \cellcolor{customgreen}15.8 & 7.3 & \cellcolor{customgreen}9.2 & \cellcolor{customgreen}10.82 & 9.9 & 23.5 & \multicolumn{1}{r}{\cellcolor{customgreen}25.6}\\

        \OURS ($\uparrow$) & 50.4 & \cellcolor{customgreen}53.3 & 28.5 & \cellcolor{customgreen}30.8 & 22.8 & \cellcolor{customgreen}25.2 & 33.2 & \cellcolor{customred}34.3 & 33.6 & \multicolumn{1}{r}{\cellcolor{customgreen}34.0}  \\
        \bottomrule
        \end{tabular}
        }
        \renewcommand{\arraystretch}{1.05}
        \setlength{\tabcolsep}{8pt}
        \centering
        \footnotesize
        \caption{\textbf{Comparison of evaluation metrics on latent diffusion models.} We compare \Ours with existing metrics on latent diffusion models. "Humans" represents the winning votes obtained by the models in the human evaluation. Each row uses a green or red cell to denote whether the ranking agrees or disagrees with human judgment respectively.
        We observe significant disagreement with human ratings for metrics, except \Ours and CLIPScore, compared to results on pixel diffusion models.}

        \label{exp:tab:comp_metrics_ldm}
        \resizebox{\linewidth}{!}{
        \begin{tabular}{@{}lcccccccccR@{}}
        \toprule
        Method & \multicolumn{2}{c}{Attribute} & \multicolumn{2}{c}{Color} & \multicolumn{2}{c}{Count} &  \multicolumn{2}{c}{Shape} & \multicolumn{2}{c}{Spatial} \\
         & \multicolumn{2}{c}{binding} &  &  &   &  \\
         \midrule
         & CLIP & T5  & CLIP  & T5 & CLIP & T5  & CLIP  & T5 & CLIP & T5  \\
         \midrule
        Humans & 14 & \cellcolor{customgreen}140 & 27 & \cellcolor{customgreen}69 & 11 & \cellcolor{customgreen}140 & 25 & \cellcolor{customgreen}48 & 21 & \multicolumn{1}{r}{\cellcolor{customgreen}73} \\
        CLIPScore ($\uparrow$) & 0.89 & \cellcolor{customgreen}0.99  & \cellcolor{customred}0.85 & 0.80 & 0.76 & \cellcolor{customgreen}0.80 & 0.76 & \cellcolor{customgreen}0.86 & 0.75 & \multicolumn{1}{r}{\cellcolor{customgreen}0.83}\\
        MID ($\downarrow$) & -8.1$\mathrm{E}$14 & \cellcolor{customgreen}-1.3$\mathrm{E}$14& -1.1$\mathrm{E}$5 & -1.1$\mathrm{E}$5 & \cellcolor{customred}-2.1$\mathrm{E}$4 & -7.6$\mathrm{E}$3 & \cellcolor{customred}-1.1$\mathrm{E}$3 & -2.1$\mathrm{E}$3 & 4.33$\mathrm{E}$14 & \multicolumn{1}{r}{\cellcolor{customgreen}-8.0$\mathrm{E}$15}\\
        VPEval ($\uparrow$)&  60.4 &\cellcolor{customgreen} 66.0 & \cellcolor{customred}87.5& 85.7 & \cellcolor{customred}64.7 & 18.9 & \cellcolor{customred}92.9 & 92.2 & \cellcolor{customred}34.1 & \multicolumn{1}{r}{23.4}\\
        LLMScore ($\uparrow$)&  \cellcolor{customred}9.2 & 9.1 & 15.4 &\cellcolor{customgreen} 17.3 & \cellcolor{customred}9.6 & 8.4 & \cellcolor{customred}11.6 & 10.6 & 21.4 & \multicolumn{1}{r}{\cellcolor{customgreen}23.1}\\

        \OURS ($\uparrow$) & 51.0 &\cellcolor{customgreen} 54.1 & 33.0 & \cellcolor{customgreen}34.4 & 24.4 &\cellcolor{customgreen}26.3 & \cellcolor{customred}33.8 & 33.5 & 33.4 &\multicolumn{1}{r}{\cellcolor{customgreen}35.8}  \\
        \bottomrule
        \end{tabular}
        }
    \end{minipage}\hfill

\end{table}

\par \noindent \textbf{Comparison with other metrics.}
In Tables~\ref{exp:tab:comp_metrics_pdm} and \ref{exp:tab:comp_metrics_ldm}, we compare \OURS with metrics that utilize external models on PDMs and LDMs, respectively.
We compare \Ours with the well-known CLIPScore\cite{hessel-etal-2021-clipscore} and recently proposed metrics such as MID~\cite{kim2022mid}, LLMScore~\cite{lu2023llmscore}, and VPEval~\cite{Cho2023VPT2I}.
In each table, we compute the scores for models using CLIP and T5 text encoders across five tasks (see Section~\ref{subsec:exp:benchmark}) and compare them with human ratings.
The row labeled ``Human'' in both tables indicates the winning votes obtained by each model.
All rows use a green or red cell to denote whether the ranking among models with CLIP and T5 text encoders agrees or disagrees with human judgment, respectively.
As mentioned in Section~\ref{sec:intro}, we observe that the value ranges of MID vary wildly, making them uninterpretable and incomparable to the rest.
CLIPScore measures the cosine similarity of the generated image with the input text prompt and has a range of $[-1, 1]$, while VPEval, LLMScore, and \Ours compute accuracy.
On the PDMs, we note that each metric disagreed with human judgment on at least one split. On the LDMs, we observe significant disagreement among different models. CLIPScore and \OURS are the only two models with
significantly less disagreement with human raters. In Fig.~\ref{exp:fig:spearman_corr}, we show the Spearman's rank correlation $\rho$~\cite{ca468a70-0be4-389a-b0b9-5dd1ff52b33f} between each metric and the human ratings.
On the x-axis, we plot $\rho$ for the PDM's results, and the LDM's results are plotted on the y-axis. The shaded region in the plot indicates the desired range for the correlations. Similar to Table~\ref{exp:tab:comp_metrics_ldm},
we observe that all metrics except \Ours have a negative correlation with human ratings. \Ours is the only metric that correlates well with both PDMs and LDMs.
Note that while all existing metrics compute the score on generated images, our proposed metric, \OURS, uses the ground truth image-text pair to compute text faithfulness.
These findings underscore the robustness of \Ours in aligning with human judgment across different models and tasks, highlighting its potential as a reliable metric for evaluating text faithfulness of generative models.

\par \noindent \textbf{Qualitative results.}
Figure~\ref{fig:qual1} (Top) compares the generations of pixel diffusion models that use T5 and CLIP text encoders. In each example, the image on the left and right are generated using CLIP and T5 text encoder respectively.
We notice that as the difficulty of prompts increases, models with a stronger text encoder performs better. Both the models fail on the much harder task of counting instances and spatial relationships.
In Figure~\ref{fig:qual1} (Bottom), each example consists two images generated using a pixel diffusion model (left) and a latent diffusion model (right) with a T5 text encoder. We observe that unlike the pixel diffusion model, the latent diffusion model can get small yet important details right (``gray table cloth'' and ``white handles'' in the second and sixth example respectively).
We believe that the latent diffusion model can offload the high frequency appearance details to the autoencoder, allowing it to pay more attention to the conditioning variable.

\begin{table}[!t]
    \begin{minipage}[t]{0.48\linewidth}
        \centering
        \vspace{0pt}
        \setlength\abovecaptionskip{0pt}
        \includegraphics[width=\textwidth]{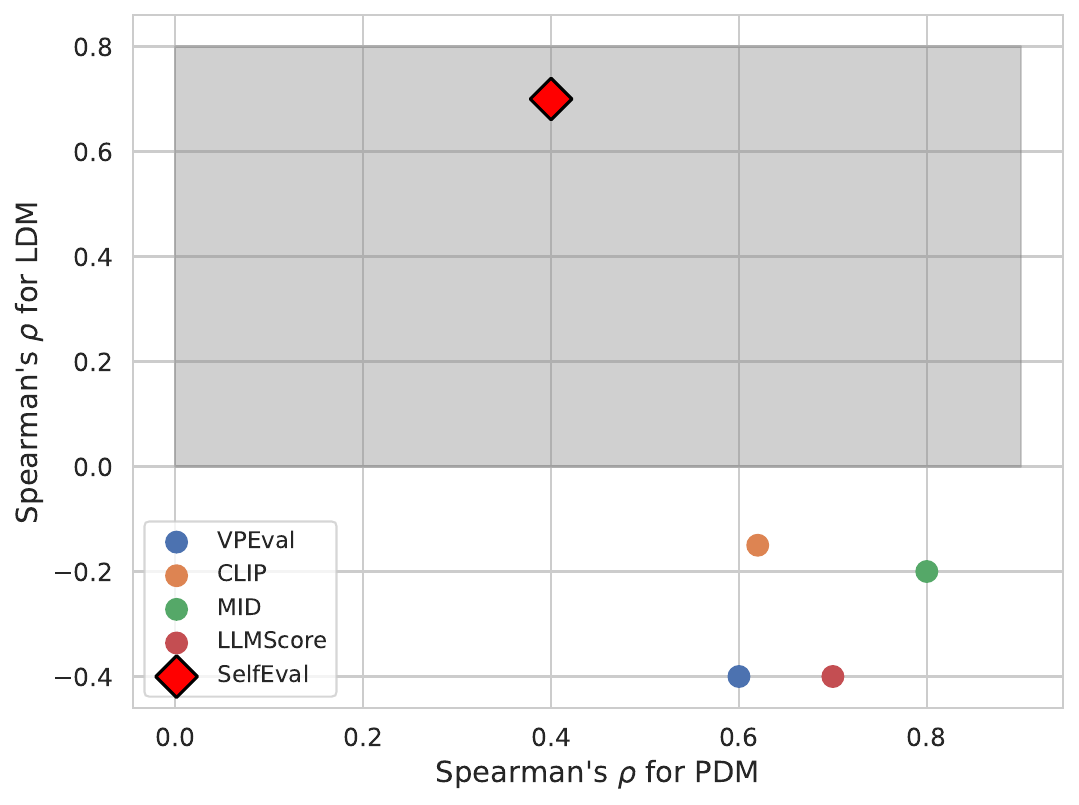}
        \captionof{figure}{\textbf{Spearman's correlation with ground truth across evaluation metrics.} We compute the Spearman rank correlation between human ratings and metrics in Tables~\ref{exp:tab:comp_metrics_pdm},~\ref{exp:tab:comp_metrics_ldm}. Existing metrics show good correlation with pixel diffusion models but a negative correlation with latent diffusion models. In contrast, \OURS positively correlates with both types of models.  }\label{exp:fig:spearman_corr}
    \end{minipage}\hfill
    \begin{minipage}[t]{0.5\linewidth}
        \renewcommand{\arraystretch}{1.0}
        \setlength{\tabcolsep}{8pt}
        \centering
        \footnotesize
        \caption{\textbf{Drawback of MID~\cite{kim2022mid}}. We compute MID using two CLIP models of two LDM-CLIP models with different backbones. Like CLIP Score, MID is sensitive to the CLIP model used for evaluation.}
        \label{exp:tab:mid_drawback}
        \resizebox{\linewidth}{!}{
            \resizebox{\linewidth}{!}{%
            \begin{tabular}{lcc}
                \toprule
              \multirow{2}{*}{Model} &  \multicolumn{2}{c}{MID$\downarrow$} \\
              \cline{2-3}
              & ViT-B/32 & ViT-L/14 \\
              \hline
              LDM-CLIP (ViT-L/14) & \textbf{27.77} & 25.70  \\
              LDM-CLIP (ViT-H/14) & 29.25& \textbf{23.53} \\
              \hline
            \end{tabular}
            }}
            \renewcommand{\arraystretch}{1.2}
            \setlength{\tabcolsep}{4pt}
            \centering
            \footnotesize
            \caption{\textbf{Performance of CLIP on the benchmark.} We evaluate the zero-shot performance of CLIP (ViT-L/14) on the six tasks. ``Random'' is the chance accuracy.
            CLIP achieves impressive performance on the tasks of \taskStyle{Color} and \taskStyle{Shape}. The performance of CLIP is close to random on the remaining tasks
            making it unsuitable for evaluating generative models on such prompts.
            }\label{exp:tab:clip_cls_acc}
            \newcolumntype{L}{>{\hspace*{-\tabcolsep}}c}
            \newcolumntype{R}{c<{\hspace*{-\tabcolsep}}}
            \resizebox{\linewidth}{!}{
            \begin{tabular}{Lccccccr}
            \toprule
            Model   &  Attribute & Color & Count & Shape &Spatial & Text \\
            & binding & & & & &corruption\\
            \midrule
            \belowrulesepcolor{Gray}
            \rowcolor{Gray}Random &  50 & 25 & 25 & 33 & 25 & 20 \\
            \aboverulesepcolor{Gray}
            \midrule
            CLIP &  55.40 & 85.20 & 67.80 & 91.10 & 40.50 & 51.00\\
            \bottomrule
            \end{tabular}
            }
        \end{minipage}

\end{table}

\subsection{Generative models applied to other reasoning tasks}\label{subsubsec:exp:abl:retrieval}

We now use the challenging Winoground~\cite{Thrush2022WinogroundPV} benchmark to evaluate the vision-language reasoning abilities of diffusion models.
Winoground defines two tasks - (1) `text score' that involves choosing the right text prompt amongst distractors given an input image; and (2) `image score' that involves picking the right image amongst distractor images given a text prompt.

\par \noindent \textbf{\OURS vs concurrent work}
Concurrent work from~\cite{li2023diffusion} demonstrates that diffusion models perform well on the Winoground text score task and achieve competitive performance with discriminative models.
Using their formulation yields poor results (zero accuracy) on the image score task as shown in Table~\ref{exp:tab:add:retr}.
\cite{li2023diffusion} use the ELBO loss as a proxy for the likelihood $p(\bx | \bc)$ which works well for comparing different text prompts and thus leads to good text score performance.
However, our analysis revealed that the ELBO loss computed for the predictions from two different images are not comparable, and thus leads to zero image score.
\OURS on the other hand, doesn't approximate the likelihood but instead estimates it as described in Sec~\ref{sec:method}.
Using \OURS leads to a non-zero image-score for the same generative model used by~\cite{li2023diffusion}, and yields performance close to that of the discriminative CLIP ViT-L model.

\par \noindent \textbf{\OURS applied to other diffusion models.}
Using \OURS reveals that all the diffusion models introduced in~\cref{subsubsec:exp:benchmark:models} achieve competitive performance on both the image score and text score tasks.
Compared to all the discriminative CLIP models, generative models achieve strong results in both image and text scores using \OURS.
This result reinforces the notion that optimizing the generative objective can provide non-trivial and complementary improvements for several visuo-linguistic reasoning tasks.
 For additional analysis on the effect of various hyperparameters on the Winoground performance, refer to the supplement.

\begin{table*}[!t]
    \begin{minipage}[t]{\linewidth}
            \vspace{0pt}
            \centering
            \includegraphics[width=\textwidth]{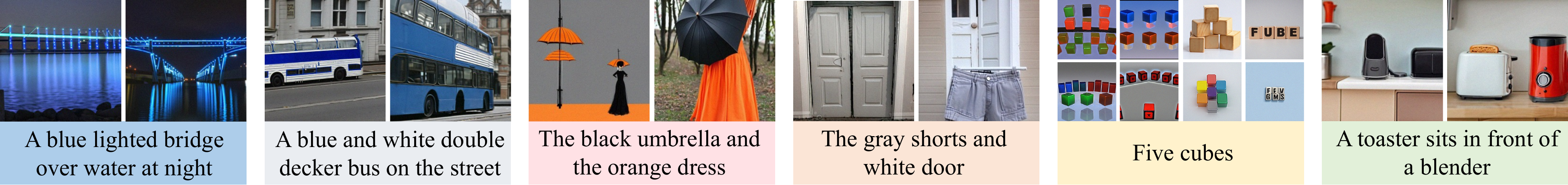}
            \centering
            \includegraphics[width=\textwidth]{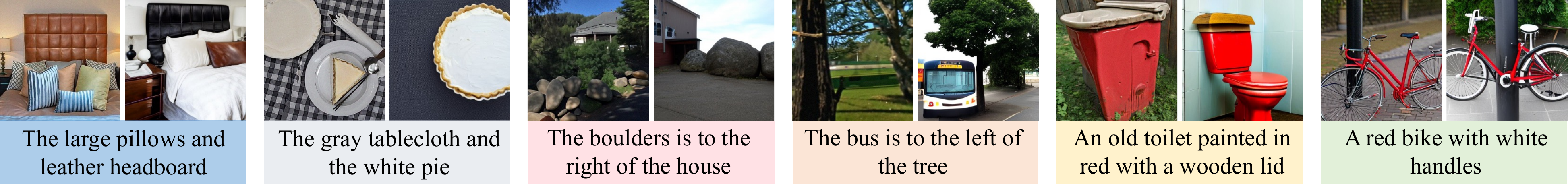}
          \captionof{figure}{\textbf{Qualitative Results.} (Top): Each example compares the generations of pixel diffusion models with CLIP (left) and T5 (right) text encoders. As the difficulty of the prompt increases, models with stronger text encoders maintain higher text fidelity.
          Both the models fail on simple prompts from \taskStyle{Count} and \taskStyle{Spatial relationships}. (Bottom): Comparison between generations of Pixel (left) and Latent (right) diffusion models with a T5 text encoder. Latent diffusion models can get smaller details like ``gray cloth'' and ``white handles'' (second and last example respectively) correctly.}\label{fig:qual1}
        \end{minipage}
\end{table*}

\subsection{Drawbacks of existing metrics}
In this section, we discuss limitations of metrics computed using external models like CLIP and LLMs that \OURS can effectively address.
CLIP score, the most common metric for evaluating text faithfulness of generative models, measures the cosine similarity between the features of the generated image and the conditioned text caption.
Recently, Mutual Information Divergence (MID)~\cite{kim2022mid} used CLIP features to compute the negative Gaussian cross-mutual information between the image and text prompt.
LLMScore~\cite{lu2023llmscore} evaluates the text faithfulness of text-to-image models by prompting an LLM to generate a score and rationale given a generated image and a text prompt.
VPEval~\cite{Cho2023VPT2I} is an interpretable and explainable metric for text-to-image generation based on visual programming~\cite{Gupta2022VisualPC}\cite{surismenon2023vipergpt}.

\par \noindent \textbf{Sensitivity to the exact CLIP model.}
We report the CLIP similarity scores of the generations from two versions of the Latent Diffusion Models~\cite{rombach2022high} on prompts from DrawBench~\cite{Saharia2022PhotorealisticTD}, Winoground~\cite{Thrush2022WinogroundPV} and COCO-minival~\cite{2014-lin} datasets in Figure~\ref{fig:clip-drawbacks}.
The first model (LDM-CLIP (ViT-L/14)) uses the text encoder of CLIP with ViT-L/14 backbone and the second model (LDM-CLIP (ViT-H/14)) uses the text encoder with OpenCLIP~\cite{ilharco_gabriel_2021_5143773} ViT-H/14 visual backbone.
Across all the three datasets, we observe that LDM-CLIP (ViT-L/14) ranks higher than LDM-CLIP (ViT-H/14) if a CLIP (ViT-L/14 visual backbone) model is used, but ranks lower with an OpenCLIP (ViT-H/14 visual backbone).
Our hypothesis is that images generated by a model using a particular CLIP text encoder may still contain some residual information, which could cause them to receive higher scores when assessed using the same CLIP model.
This type of bias was identified by~\cite{park2021benchmark} in the context of evaluation of text-to-image models, though not in relation to the CLIP score.
We observe that this bias is not limited to the CLIP score but pertains to any metric that uses CLIP features.

Table~\ref{exp:tab:mid_drawback} shows the MID~\cite{kim2022mid} computed using CLIP ViT-B/32 and ViT-L/14 backbones of two models, LDM-CLIP (ViT-L/14) and (ViT-H/14) on prompts from the Winoground~\cite{Thrush2022WinogroundPV} dataset.
Our observations show that when MID is computed using ViT-B/32, LDM-CLIP (ViT-H/14) ranks higher than LDM-CLIP (ViT-L/14).
Conversely, when MID is computed using the ViT-L/14 backbone, LDM-CLIP (ViT-L/14) ranks higher than LDM-CLIP (ViT-H/14).
We emphasize the need for caution among researchers who employ this metric, particularly concerning this bias.
\OURS avoids this problem as we do not employ an external model for evaluation.

\par \noindent \textbf{CLIP score is limited by CLIP's performance} and thus using it as a proxy on tasks where CLIP itself has poor performance does not yield meaningful comparsions.
While the CLIP model has demonstrated impressive zero-shot performance on several image-text tasks, it has severe limitations on many complex reasoning tasks.
We compute the performance of CLIP ViT-L/14 model on the six tasks introduced in~\cref{subsec:exp:benchmark} and report the results in~\cref{exp:tab:clip_cls_acc}.
CLIP performs well on \taskStyle{Color} and \taskStyle{Shape} but its performance on all the other tasks is poor.
On the widely used DrawBench prompts, 25\% of the captions evaluate the generations for attribute binding, counting, spatial relationships and text corruption.
Thus, using CLIP to evaluate generations on such prompts in DrawBench is not ideal.
\OURS avoids this problem by directly leveraging the diffusion model itself.

\begin{table*}[!t]
    \begin{minipage}[t]{\linewidth}
            \vspace{0pt}
            \centering
            \includegraphics[width=\textwidth]{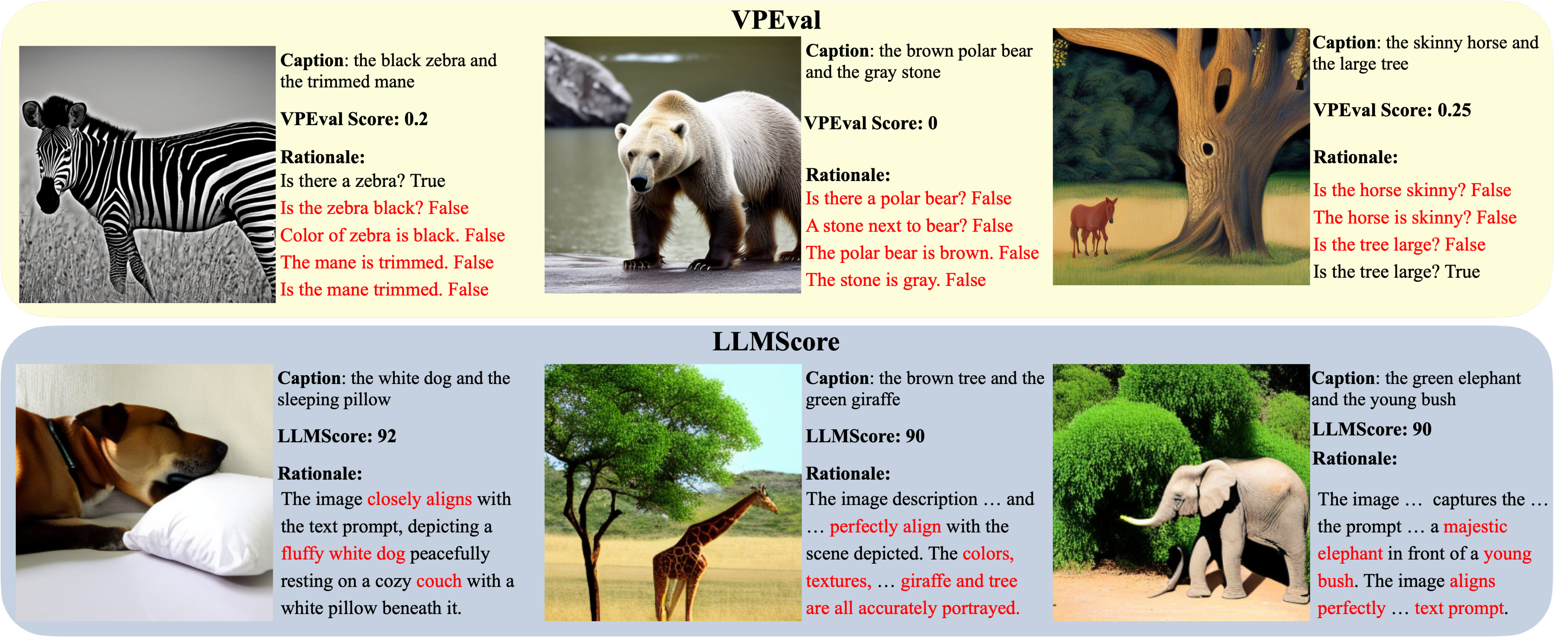}
          \captionof{figure}{\textbf{Hallucination} in large language and vision models affects evaluation of generative models. (Top): Three examples showing the caption, scores and corresponding rationale from VPEval~\cite{Cho2023VPT2I}.
          Wrong entries are highlighted in red. We observe that VPEval penalizes the generation despite its faithfulness to the text, due to hallucinations of the vision and language model. (Bottom): Three examples showing the generated image,
          a wrong caption and the rationale generated by LLMScore~\cite{lu2023llmscore}. LLMScore assigns a high score despite the incorrect caption, due to the hallucinations of the LLM, as highlighted in red in the rationale.
 }\label{exp:fig:llm_drawback}
        \end{minipage}
\end{table*}

\par \noindent \textbf{Hallucination in LLMs affects evaluation}. VPEval~\cite{Cho2023VPT2I} and LLMScore~\cite{lu2023llmscore} employ vision foundation models to ground the concepts, mentioned in the text prompt, in the generated image and
leverage a LLM to verify the prompt adherence. This would work if the LLMs are robust, but as shown in Fig.~\ref{exp:fig:llm_drawback}, LLMs hallucinate irrelevant information making the evaluation unreliable. This is reflected in the worse
than chance performance of LLMScore~\cite{lu2023llmscore} in Tables~\ref{exp:tab:comp_metrics_pdm},~\ref{exp:tab:comp_metrics_ldm} across all the metrics making them unsuitable for such evaluation tasks. We reiterate that the performance of the external model on the evaluation set has a
large impact on its evaluation capabilities and \Ours is one such way to eliminate the reliance on external models for evaluation.

\section{Conclusion}
\label{sec:conclusion}
This paper introduced \OURS, an automated method for evaluating the text-understanding capabilities of diffusion models.
\OURS estimates the likelihood of real images given text prompts using the diffusion model itself, eliminating the need for external discriminative models.
Our experiments demonstrated that \OURS aligns with human evaluations across various models and tasks, proving its reliability as an automated metric for text conditioned image generation.
We anticipate that such metrics will expedite diffusion model research and encourage further improvements.
\OURS's applicability extends beyond text-to-image diffusion models, potentially serving in the evaluation of other conditioned diffusion models like text-to-audio and text-to-video.
Future work aims to generalize \OURS for use with non-diffusion based generative models.

\section{Broader Impact}
This research introduces \OURS, an automatic evaluation method, eliminates the use of an external model for evaluation while demonstrating strong agreement with
human annotators.
This automated evaluation could significantly reduce the monetary and time costs associated with hiring human evaluators, making the process of developing and refining generative models more efficient.
Progress in research areas related to \OURS could potentially improve the ability of generative models to adhere to text.
However, such models pose a serious concern as they could lead to the creation of deceptive images if placed in the wrong hands.

\bibliography{references}
\bibliographystyle{tmlr}

\supplementarytitle
\clearpage
\appendix
\section{Additional Details of \OURS}

\begin{figure}[!h]
  \centering
  \includegraphics[width=\textwidth]{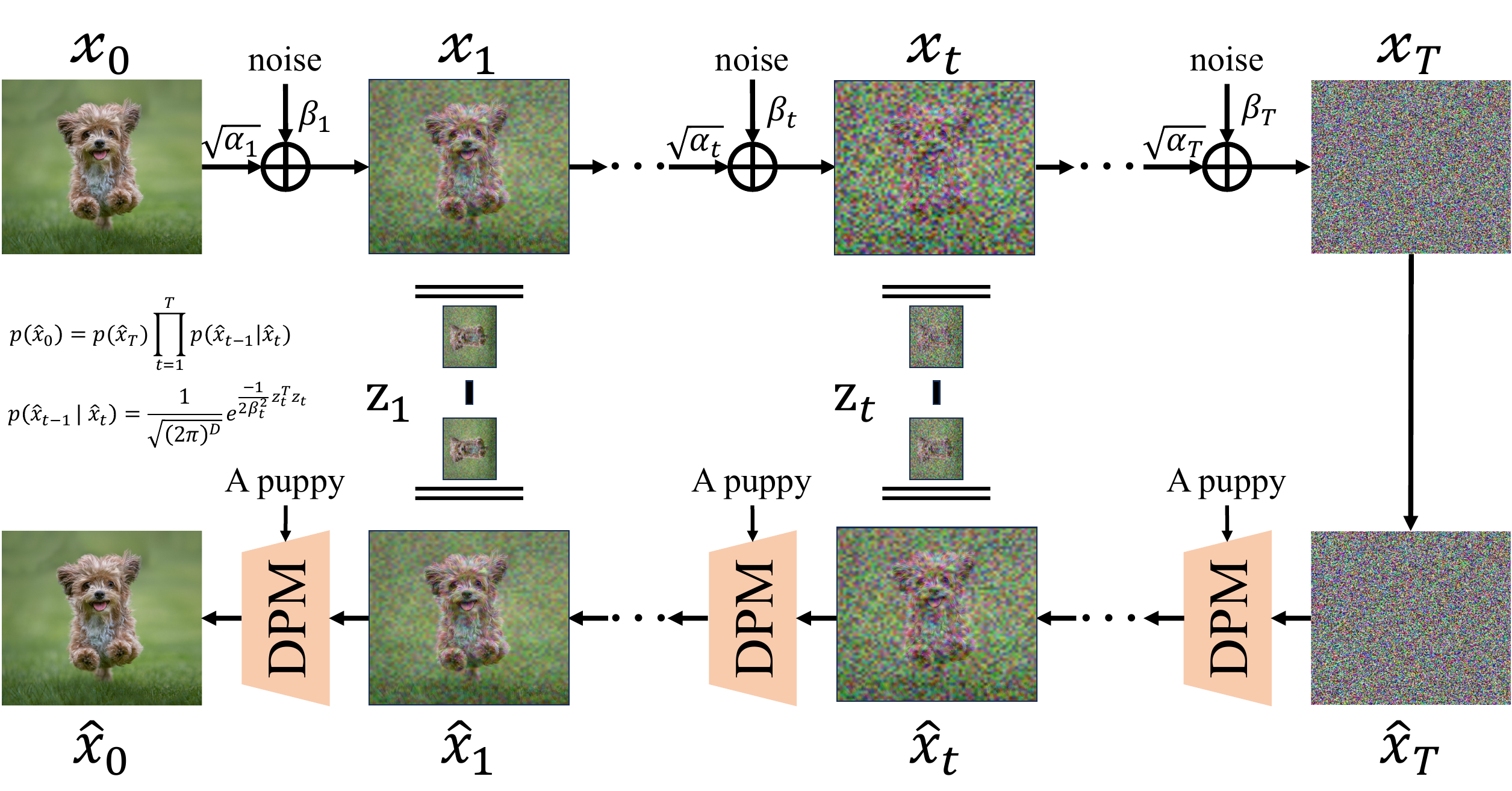}
\caption{\textbf{Illustration of proposed method:} (Left) Starting from a noised input, the standard diffusion sampling method denoises the input iteratively to generate images from the input distribution.
(Middle): SelfEval takes an image $x_0$ and conditioning $c$ pairs to estimates the likelihood $p(x_0 | c)$ of the pair in an iterative fashion. (Right): Given an image, $x_0$ and $n$ captions, $\{c_0, c_1, \dotsc, c_n\}$, SelfEval is a principled way to convert generative models into discriminative models.
In this work, we show that the classification performance of these classifiers can be used to evaluate the generative capabilities.}\label{sup:fig:main_fig}
\end{figure}

In this section, we provide a detailed algorithm and systematic figure of \OURS in Algorithm~\ref{alg:full} and Figure~\ref{sup:fig:main_fig} respectively. \OURS iteratively denoises an image, similar to the reverse process of diffusion models, but instead estimates the likelihood of an image-text pair.

\begin{algorithm}[!h]
  \begin{algorithmic}[1]
  \STATE \textbf{Input}: Diffusion model $p_\theta (\bx_{t-1}|\bx_t)$; Input image $\bx_0$; Forward latents: $\{\bx_{1:T}\}$; Reverse latents: $\{\hat{x}_{1:T}\}$; Number of trials: $N$

  \FOR {i=1:$N$}

  \STATE Sample $\text{noise}\sim \mathcal{N}(0, \mathbb{I}) $
  \STATE $\bx_{1:T} = q_{\text{sample}}(\bx_0, t=1:T, \text{noise}=\text{noise})$; $\bx_t \in \mathbb{R}^D$
  \STATE conditionals $\longleftarrow$ [ ]
  \FOR{j=1:T}
  \STATE $p(\bx_{t-1} | \bar{\bx}_t, \bc)$ = $\frac{1}{\sqrt{(2\pi)^{D}|\bSigma_\theta|}}e^{-0.5(\bx_{t-1} - \bmu_{\theta}(\bar{\bx}_t, t, \bc))^T \bSigma_\theta^{-1}(\bx_{t-1} - \bmu_{\theta}(\bar{\bx}_t, t, \bc))}$
  \STATE conditionals = [conditionals ; $p(\bx_{t-1} | \bar{\bx}_t, \bc)$]
  \ENDFOR
  \STATE Compute $p(\bx_T) = \frac{1}{\sqrt{(2\pi)^{D}}}e^{\frac{-1}{2\beta_T^2}\|\bx_{T}\|^2}$
  \STATE Compute likelihood $p_i(\bx_0|\bc) = p(\bx_T)\prod_{t=1}^{T}p(\bx_{t-1} | \bar{\bx}_t,\bc)$

  \ENDFOR
  \STATE $p(\bc|\bx_0) = \frac{p(\bx_0 | \bc)}{|\mathcal{C}|}$
  \end{algorithmic}
  \caption{Algorithm for estimating $p(\bx|\bc)$ using \OURS}
  \label{alg:full}
  \end{algorithm}

  \section{Details of Human evaluation}
\begin{figure}[!h]
  \centering
 \includegraphics[width=\textwidth]{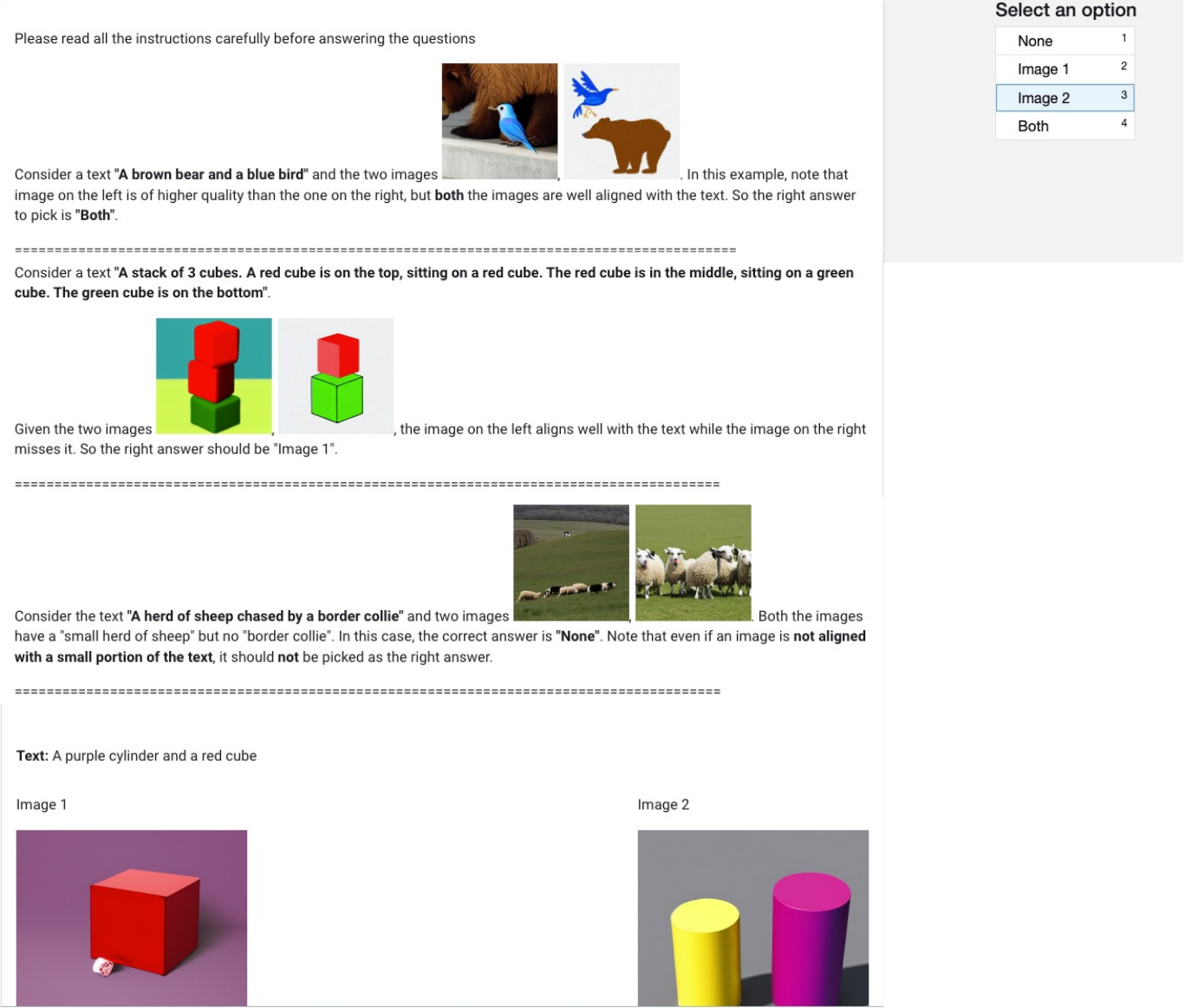}
 \caption{\textbf{Template for Human raters.} The template consists of instructions explaining the nature of the task (top) followed by a text prompt with two generations (bottom). Humans are expected to pick one of four options (shown on the right): ``both'' the generations are faithful, ``none'' of them are faithful, or if only one of the two images (``Image 1'' or ``Image 2'') demonstrates fidelity to the text prompt.}\label{sup:fig:he_template}
\end{figure}

\begin{figure}[!h]
\centering
\includegraphics[width=\textwidth]{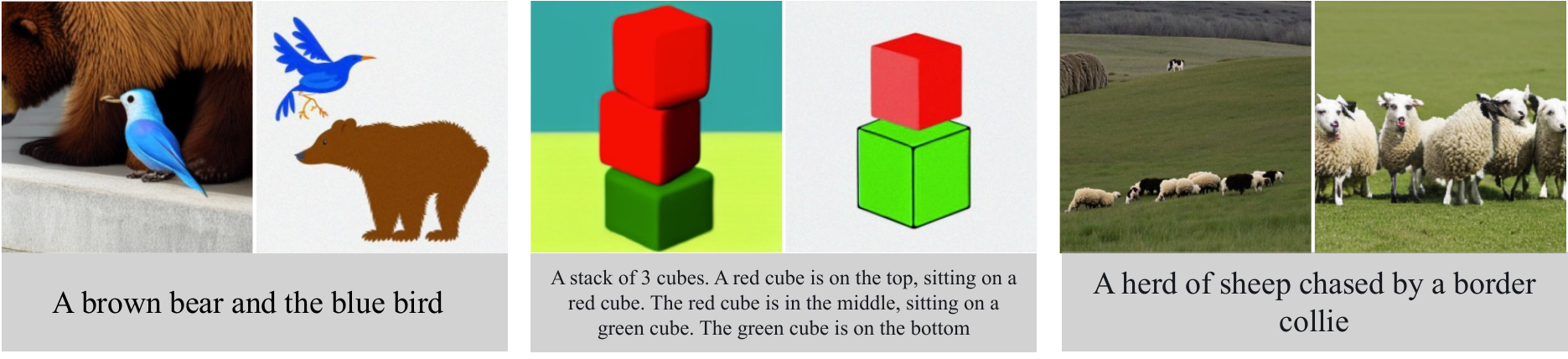}
\caption{\textbf{Instructions for Human raters.} We provide three examples describing all the possible scenarios. The first example shows two images that are faithful to the text but with varying image qualities. To prevent the raters from conflating image quality with text faithfulness, we recommend the raters to pick ``both'' for such examples. The second example illustrates a case where only one of the image is faithful to the text. In this case, the raters are adviced to pick
the option corresponding to the right image (``Image 1'' in this case). The final example shows a case where both the examples are not faithful to the text (there is no border collie), in which case, we advice the raters to pick ``none''. }\label[type]{sup:fig:he_instruction}
\end{figure}

Human evaluations are the de-facto standard for judging the performance of text-to-image models.
we adopt a conventional A/B testing approach, wherein raters are presented with generations from two models and are asked to vote for one of four choices: ``both'' the generations are faithful, ``none'' of them are faithful, or if only one of the two models (``model 1'' or ``model 2'') demonstrates fidelity to the given prompt.
We show the template provided to the raters in Figure~\ref{sup:fig:he_template}.
The template includes three examples that advice the raters on how to rate a given sample followed by a text prompt and two images. The four possible choices are shown on the right in Figure~\ref{sup:fig:he_instruction}.
The images used as instructions for the human raters are shown in Figure~\ref{sup:fig:he_instruction}. Figure~\ref{sup:fig:he_instruction} shows three pairs of images with the text prompt below them.
The first example shows two images that are faithful to the input prompt but the quality of one (left) image superior to the other (right). Since, we ask the raters to evaluate the text faithfulness, we recommend picking the ``both'' option for such samples.
The second image shows an example where only one of the images is faithful to the text. The raters are instructed to pick the option corresponding to the right image in this case.
The final example shows two images that are not faithful to the text prompt. The raters are adviced to pick the ``none'' option in this scenario.

\section{Ablation Experiments}
\begin{table}[!h]
  \begin{minipage}[b]{0.34\linewidth}
  \renewcommand{\arraystretch}{1.0}
  \setlength{\tabcolsep}{4pt}
  \centering
  \footnotesize
  \caption{\textbf{Effect of timesteps} on the performance of \OURS on the six splits.}
  \label{sup:tab:split_abl_T}
   \resizebox{\linewidth}{!}{
  \begin{tabular}{@{}lcccccc@{}}
  \toprule
T &  Attribute & Color & Count & Shape & Spatial & Text \\
& & & & & & Corruption \\
   \midrule
   50&  54.2 & 32.2 & 26.3 & 34.9 & 33.0 & 25 \\
   100&  54.3 & 34  & 25.8 & 30.2 & 38.0 & 24.3\\
   250&  53 & 32.3  & 27.4 & 35 & 32.7 & 21.7\\
  \bottomrule
  \end{tabular}
   }
  \end{minipage}
  \begin{minipage}[b]{0.33\linewidth}
    \renewcommand{\arraystretch}{1.0}
    \setlength{\tabcolsep}{8pt}
    \centering
    \footnotesize
    \caption{\textbf{Effect of N} on the performance of \OURS on the six splits.}
    \label{sup:tab:split_abl_N}
     \resizebox{\linewidth}{!}{
    \begin{tabular}{@{}lcccccc@{}}
    \toprule
  N &  Attribute & Color & Count & Shape & Spatial & Text \\
  & & & & & & Corruption \\
     \midrule
     1&  53.0 & 26.0 & 27.2 & 35.2 & 31.2 & 20.7 \\
     5&  54.3 & 31.7  & 25.7 & 34.9 & 33.0 & 22.1\\
     10&  54.3 & 34.0  & 25.8 & 32.5 & 38.6 & 24.3\\
     15&  53.4 & 36.3  & 28.0 & 36.3 & 32.8 & 22.8\\

    \bottomrule
    \end{tabular}
     }
    \end{minipage}
    \begin{minipage}[b]{0.33\linewidth}
      \renewcommand{\arraystretch}{1.0}
      \setlength{\tabcolsep}{8pt}
      \centering
      \footnotesize
      \caption{\textbf{Effect of the choice of seed} on the performance of \OURS.}
      \label{sup:tab:split_abl_seed}
       \resizebox{\linewidth}{!}{
      \begin{tabular}{@{}lcccccc@{}}
      \toprule
    S &  Attribute & Color & Count & Shape & Spatial & Text \\
    & & & & & & Corruption \\
       \midrule
       1&  54.3 & 34.0  & 25.8 & 32.5 & 38.6 & 24.3\\
       2&  53.0 & 26.0 & 27.2 & 35.2 & 31.2 & 20.7 \\
       3&  54.3 & 31.70  & 25.7 & 34.9 & 33.0 & 22.1\\
       std&  0.5 & 0.5  & 0.9 & 1.4 & 1.5 & 0.8\\

      \bottomrule
      \end{tabular}
       }
      \end{minipage}
\end{table}
In this section we analyze the effect of various components that affect the performance of \OURS on the six splits introduced in the main paper.
We use the LDM-T5 model for all our experiments.

\noindent\textbf{Effect of T}: \OURS has a time complexity of $\mathcal{O}(NT)$ and Table~\ref{sup:tab:split_abl_T} shows the the effect of timesteps on the performance of \OURS.
We observe that \OURS achieves the best result at different timesteps for different datasets. We notice that the performance drops as we increase the timesteps from $100$ to $250$ in most cases.
As the number of timesteps increases, we believe that the fraction of them responsible for text faithfulness decrease, resulting in a drop in performance.
We find $T=100$ to be a good tradeoff for performance and speed and is used for all the experiments on the six data splits.

\noindent\textbf{Effect of N}: Table~\ref{sup:tab:split_abl_N} shows the results of the effect of number of trials $N$ on the performance of \OURS.
We observe that $N=10$ works best across all the six splits and is the default choice for $N$.

\noindent\textbf{Effect of seeds}: \OURS corrupts an input image using standard gaussian noise in each trial and we analyze the effect of the seed on the performance of \OURS in Table~\ref{sup:tab:split_abl_seed}.
We observe that the performance is stable across all the six splits with a standard deviation within 1 percentage point in most of the cases.
We report the seed number instead of the actual value for brevity and use the seed 1 as the default choice for all the experiments.

\section{Additional experiments on Winoground}
In this section we ablate a few design decisions on the Winoground dataset.
We use the LDM-T5 model for all the experiments.
\begin{table}[!h]
  \begin{minipage}[t]{0.32\linewidth}

  \renewcommand{\arraystretch}{1.0}
  \setlength{\tabcolsep}{4pt}
  \centering
  \footnotesize
  \caption{\textbf{Effect of timesteps} on the performance of \OURS on the Winoground dataset}
  \label{sup:tab:wg_abl_T}
   \resizebox{\linewidth}{!}{
  \begin{tabular}{@{}lcc@{}}
  \toprule
T &  Image Score & Text Score \\
   \midrule
  20 & 11.50 &  30.75 \\
   50&  13.50 & 29.00 \\
   100&  12.25 & 25.25 \\
   250&  11.25 & 27.75 \\

  \bottomrule
  \end{tabular}
   }
\end{minipage}
\begin{minipage}[t]{0.32\linewidth}
  \renewcommand{\arraystretch}{1.0}
  \setlength{\tabcolsep}{5pt}
  \centering
  \footnotesize
  \caption{\textbf{Effect of the \# of trials} on the performance of \OURS on the Winoground dataset}
  \label{sup:tab:wg_abl_N}
   \resizebox{\linewidth}{!}{
  \begin{tabular}{@{}lcc@{}}
  \toprule
  N &  Image Score & Text Score \\
   \midrule
  1 & 17.00 &  26.25 \\
   5&  14.75 & 26.00 \\
   10&  13.50 & 29.00 \\
   20&  11.25 & 24.75 \\

  \bottomrule
  \end{tabular}
   }
\end{minipage}
\begin{minipage}[t]{0.34\linewidth}
  \renewcommand{\arraystretch}{1.0}
  \setlength{\tabcolsep}{7pt}
  \centering
  \footnotesize
  \caption{\textbf{Effect of the choice of seed} on the performance of \OURS on the Winoground dataset}
  \label{sup:tab:wg_abl_seed}
  \resizebox{\linewidth}{!}{
  \begin{tabular}{@{}lcc@{}}
  \toprule
  S &  Image Score & Text Score \\
   \midrule
  1 & 13.50 &  29.00 \\
   2&  13.00 & 27.00 \\
   3&  12.00 & 28.50 \\
    &  12.83$\pm$ 0.76 & 28.17$\pm$1.04 \\

  \bottomrule
  \end{tabular}
  }
\end{minipage}
\end{table}

\noindent\textbf{Effect of T}: We show the effect of the number of timesteps on the performance of \Ours on the Winoground dataset in Table~\ref{sup:tab:wg_abl_T}. From Table~\ref{sup:tab:wg_abl_T}, we observe that \OURS achieves the best result for image and text score at different time steps.
Image score is a harder task compared to Text score~\cite{Thrush2022WinogroundPV} and hence \OURS needs more timesteps to perform better on Image score.
As the number of timesteps increase, we observe a drop in both Image and Text scores.
Studies~\cite{Li2023GLIGENOG} show that the earlier timesteps generate low frequency information (responsible for text fidelity), while latter ones are responsible for high frequency appearance details.
By increasing the number of timesteps, the fraction of timesteps contributing to improving the faithfulness to text (and thereby image and text scores) decreases, resulting in a drop in performance.
All other experiments on Winoground use T=50 unless otherwise specified.

\noindent\textbf{Effect of N}: We show the effect of the number of trials (N) in Table~\ref{sup:tab:wg_abl_N}.
With fewer trials, the estimates are not reliable and larger trials make it computationally expensive.
We observe that we attain a good tradeoff for performance and speed with $N=10$.

\noindent\textbf{Effect of the seed}: We show the effect of seed on the performance of \OURS in Table~\ref{sup:tab:wg_abl_seed}.
We just report the seed number for brevity. We observe that both the scores are relatively stable across different values of seed. We fix seed \#1 for all the experiments in this work.

\section[short]{Converting COCO image-caption pairs for ITM}
We use image-caption pairs from COCO for the tasks of \taskStyle{Color}, \taskStyle{Count} and \taskStyle{Spatial relationships}. We use the question answering data collected by authors of TIFA~\cite{hu2023tifa}
to construct data for our tasks. We pick only samples constructed from COCO. Given question answering samples from TIFA,
we identify the corresponding image-caption pair from COCO and replace the correct answer in the caption with the multiple choices to form samples for the task of Image-Text Matching.

\section[short]{Limitations}

SelfEval relies on the sampling of the generative model to compute the scores.
So the limitations of the sampling process of a generative model affect SelfEval.
To be precise, for a model with $T$ diffusion time steps and a classification task with $C$ classes, \OURS samples $N$ noise signals.
This results in an overall complexity of the order $\bigO(NCT)$ for computing probabilities using \Ours.
The complexity increases linearly with the number of classes $C$ making it
difficult to scale to thousands of classes (like ImageNet~\cite{5206848}).
However, several optimizations like randomly picking a starting timestep to denoise (instead of all $T$ diffusion timesteps) and efficient classification tricks~\cite{li2023diffusion}
can be employed to improve the time complexity of \Ours.
Additionally, unlike other black-box evaluation methods, which only require the generations from the model, \Ours requires the model definition and its checkpoints for evaluation making it
impossible to evaluate closed-source generative models without model definition and checkpoint.

\end{document}